\documentclass[letterpaper]{article} 
\usepackage{aaai2026}  
\usepackage{times}  
\usepackage{helvet}  
\usepackage{courier}  
\usepackage[hyphens]{url}  
\usepackage{graphicx} 
\usepackage{natbib}  
\usepackage{caption} 
\usepackage{amsmath}
\usepackage{amssymb}
\usepackage{multirow}
\usepackage{makecell}
\usepackage{tabularx}
\usepackage{graphicx}
\usepackage{algorithm}
\usepackage{algorithmic}
\usepackage{subfigure}
\usepackage{amsthm}

\usepackage{newfloat}
\usepackage{listings}
\DeclareCaptionStyle{ruled}{labelfont=normalfont,labelsep=colon,strut=off} 
\floatstyle{ruled}
\newfloat{listing}{tb}{lst}{}
\floatname{listing}{Listing}
\newtheorem{theorem}{Theorem}
\newtheorem{proposition}{Proposition}
\newtheorem{lemma}{Lemma}
\newtheorem{corollary}{Corollary}
\newtheorem{definition}{Definition}
\newtheorem{assumption}{Assumption}
\newtheorem{remark}{Remark}

\title{Bootstrapping LLMs via Preference-Based Policy Optimization}
\author {
    Chen Jia
}
\affiliations{
    SI-TECH Information Technology \\
    jiachenwestlake@gmail.com
}

\usepackage{bibentry}

\begin{document}

\maketitle

\begin{abstract}
	Bootstrapping large language models (LLMs) via preference-based policy optimization enables aligning model behavior with human preferences while reducing reliance on extensive manual annotations. We propose a novel preference-based policy optimization (PbPO) framework that formulates learning as a min-max game between the LLM policy and a reward model (RM). The RM is constrained within a confidence set derived from collected preferences to ensure reliable exploitation, while simultaneously promoting robust exploration. Our iterative online algorithm actively collects new preference data from the evolving policy, enabling continual self-improvement of both the policy and the RM. We provide theoretical guarantees, establishing high-probability regret bounds for both sequence-level and token-level RMs. Extensive experiments across five benchmark datasets demonstrate that PbPO consistently outperforms state-of-the-art preference optimization methods.
\end{abstract}

\begin{links}
  \link{Extended version}{https://arxiv.org/abs/2511.12867}
\end{links}
\section{Introduction}
Reinforcement Learning from Human Feedback (RLHF) \cite{christiano2017deep, ziegler2019fine,zhan2023provable,zhanprovable} has become a key paradigm for aligning machine learning models with human preferences \cite{stiennon2020learning, ouyang2022training, rame2024warm, rafailov2024direct}, particularly for Large Language Models (LLMs) \cite{brown2020language, touvron2023llama, achiam2023gpt}. By leveraging human annotations, RLHF trains a reward model (RM) that guides the policy to generate outputs that are helpful, truthful, and harmless \cite{ouyang2022training, bai2022training, casper2023open}.

\begin{figure}[t!]
	\centering
	\includegraphics[width=1.\linewidth]{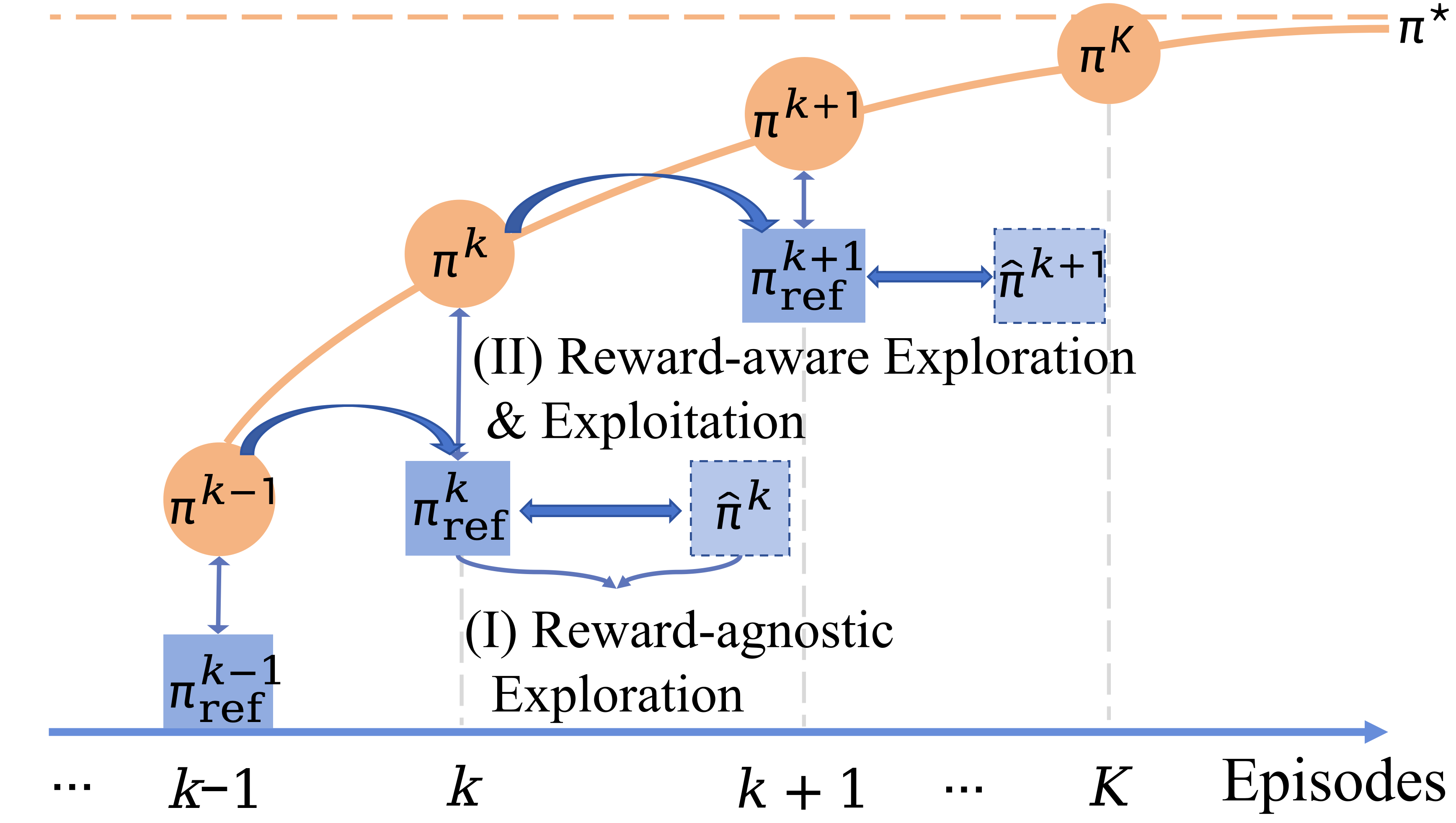}
	\caption{PbPO framework for bootstrapping LLMs. At each episode $k \in {1,2,\ldots,K}$: (I) \textbf{Reward-agnostic exploration}: collect new preference data using the current reference policy $\pi^k_{\rm ref}$ and an exploration-enhancing policy $\hat{\pi}^k$. (II) \textbf{Reward-aware exploration \& exploitation}: update the main LLM policy $\pi^k$ via a min-max objective using the reward model trained on collected preferences.}
	\label{fig:overall}
\end{figure}

Existing RLHF methods primarily rely on fixed preference datasets, separating preference data collection and RM pretraining \cite{ziegler2019fine,stiennon2020learning,ouyang2022training}. Collecting large-scale, high-quality preference data is costly, often requiring human annotators \cite{bai2022training} or LLMs \cite{cui2023ultrafeedback}. Moreover, reward misspecification and misgeneralization \cite{hongsensitivity, gao2023scaling} can lead to suboptimal policy updates.

We adopt a bootstrapping approach in which LLMs iteratively improve themselves. Following online iterative RLHF, responses are sampled from the current LLM policy, feedback is collected to generate new preference data, and the RM is updated accordingly. Recent efforts \cite{zhanprovable, xiong2024iterative, ye2024online,das2025active} enhance this loop by incorporating reward-agnostic exploration strategies that maintain uncertainty of preference data collection, thereby improving data diversity to cover the whole preference data space. But standard online RLHF frameworks often optimize the RM solely for observed preferences, risking overfitting and premature convergence.

To address this, we propose \textbf{preference-based policy optimization (PbPO)}, a unified framework integrating reward-agnostic and reward-aware exploration. PbPO formulates a min-max game between the LLM policy and the RM. The RM is constrained within a confidence set derived from collected preferences to ensure reliable exploitation, while being optimized to minimize the performance gap with a reference policy, thereby promoting robust exploration. Iteratively collecting new preference data from the evolving policy enables effective bootstrapping of LLM alignment (Figure~\ref{fig:overall}).

We provide theoretical guarantees for PbPO, establishing nearly optimal regret bounds for both sequence-level \cite{ouyang2022training, bahetileftover2024, xiong2024iterative} and token-level RMs \cite{zeng2024token, cen2025value, zhong2025dpo}, as illustrated in Table \ref{tbl:maintho}. Extensive experiments on five benchmark datasets show that PbPO consistently outperforms existing state-of-the-art methods.

Our main contributions are summarized as follows:
\begin{enumerate}
	\item We introduce PbPO, a unified framework combining reward-agnostic and reward-aware exploration within online RLHF.
	\item We provide theoretical guarantees for iterative LLM self-improvement under PbPO, covering both sequence-level and token-level RMs.
	\item We empirically validate PbPO on five benchmark datasets, achieving state-of-the-art performance.
\end{enumerate}

\section{Problem Formulation}
We consider language modeling as a sequence-to-sequence decision-making task, where an input prompt $x$ is first sampled, followed by generation of an output sequence $y = \{y_1, \ldots, y_H\}$. We formulate this process as an episodic \textit{finite-horizon} Markov Decision Process (MDP), denoted by $\mathcal{M} = (\mathcal{S}, \mathcal{A}, r^\star, \mu, H)$, where $\mathcal{S} = \{\{s_h\}_{1 \le h \le H}\}$ denotes the state space consisting of the initial prompt state $\{s_1 = x\}$ and output prefixes $\{\{s_h = (x, y_{\le h-1})\}_{2 \le h \le H}\}$, $\mathcal{A} = \{\{a_h\}_{1 \le h \le H}\}$ denotes the action space corresponding to predicted tokens $a_h = y_h$, $\mu \in \Delta_\mathcal{S}$ is the initial state distribution, and $r^\star: \tau \mapsto r^\star(\tau) \in \mathbb{R}$ denotes the ground-truth reward model for a trajectory $\tau = (s_1, a_1, \ldots, s_H, a_H) \in (\mathcal{S} \times \mathcal{A})^H$. Note that $r^\star$ is defined as a trajectory-wise reward, which is more general than the sequence-level and token-level rewards commonly used in RLHF. These commonly used rewards can be viewed as special cases and will be discussed in the following sections. Under our general MDP formulation, each of these cases corresponds to a specific sub-problem induced by a particular MDP structure.
Additionally, we assume a deterministic state-transition function in sequence generation, where each action deterministically advances the state. 

We focus on optimizing a LLM policy for predicting the next action at a state $s_h$ with a probability distribution $\pi(\cdot \mid s_h) \in \Delta_{\mathcal{A}}$. Starting from an initial state $s_1 = x \sim \mu$, a trajectory $\tau=(s_1, a_1, \ldots, s_H, a_H)$ is generated with the probability distribution $\mathbb{P}^{\pi}(\tau) = \mu(x) \cdot \big( \prod_{h=1}^{H-1} \pi(a_h \mid s_h)) \cdot \pi(a_H \mid s_H)$. Given a LLM policy $\pi$, we define its performance as the expected cumulative reward over the episode as $J(\pi, r) := \mathbb{E}_{\tau \sim \pi} \left[ r(\tau) \right]$, where $\mathbb{E}_{\tau \sim \pi}[\cdot]$ denotes the expectation w.r.t. $\tau$ drawn from the probabiliy distribution $\mathbb{P}^{\pi}(\tau)$. 

\noindent\textbf{Optimization objective.} We consider an episodic learning objective, where the learning algorithm aims to minimize the following cumulative regret over $K$ episodes:
\begin{align} \label{eq:regret}
	{\rm Regret}(K) := \sum_{k=1}^K \left(  J(\pi^\star, r^\star) -J(\pi^k, r^\star) \right),
\end{align}
where the optimal LLM policy is defined as $\pi^\star = \mathop{\arg\max}_{\pi \in \Pi} J(\pi, r^\star)$.

\noindent\textbf{Preference optimization.} 
Standard preference alignment approaches for LLMs \cite{ouyang2022training,zhu2023principled} assume ground-truth RM $r^\star$, such that the trajectory-based preference probability over a binary feedback $o \in \{0,1\}$ for a trajectory pair in precollected preference dataset $(\tau^0, \tau^1) \in \mathcal{D}^{\rm pref}$ is formulated by:
\begin{align}
	\begin{split}
		\mathbb{P}_{r^\star}(\tau^0 \succ \tau^1) =& \mathbb{P}_{r^\star}(o = 1 \mid \tau^0, \tau^1) \\
		=& \sigma \left( r^\star(\tau^0) - r^\star(\tau^1) \right),
	\end{split}
\end{align}
where $\sigma(x) = 1/(1 + \exp(-x))$ leads to the Bradley-Terry-Luce (BTL) model \cite{christiano2017deep}.  

Classical preference optimization estimate the RM $r \in \mathcal{G}_r$ from precollected preference data to approximate $r^\star$. Our analysis will follow the assumption that the reward difference are bounded in the interval $[r_{\rm min}, r_{\rm max}]$ in $\mathbb{R}$ and  leverage the quantity $\sup_{r_{\rm min} \leq  x \leq r_{\rm max}} \vert 1/\sigma'(x) \vert \leq \kappa$ with some constant $\kappa\geq0$. 

Most previous work on preference optimization for LLMs focuses on the offline setting, where preference data are drawn from a fixed, predetermined distribution \cite{ziegler2019fine,stiennon2020learning,ouyang2022training,bai2022training}. We extend this approach iteratively by updating the reference policy at each episode to be the policy optimized in the previous step, generating preference samples from this evolving policy while incorporating exploration strategies.  

\noindent\textbf{Notations.} 
We use $\mathbb{E}_{\tau^0 \sim \pi_0, \tau^1 \sim \pi_1}[\cdot]$ to denote the expectation w.r.t. first drawing a prompt $x \sim \mu$, then draws $\tau^0 \sim \mathbb{P}^{\pi_0}(\cdot \mid x)$ and $\tau^1 \sim \mathbb{P}^{\pi_1}(\cdot \mid x)$. 
We use $\| z \|_{\Sigma}$ to denote the induced norm $\sqrt{z^\top \Sigma z}$ for some positive-definite matrix $\Sigma$. We write $\widetilde{\mathcal{O}}(\cdot)$ to omit logarithmic factors and constants in regret bound.  

\begin{table}[t]
	\setlength{\tabcolsep}{3pt}
	\centering
\begin{tabular}{lcc}
	\hline
	 {\bf ${\rm Regret}(K)$}& {\bf Sequence-level RM} & {\bf Token-level RM} \\
	\hline
	 Upper Bound&$\widetilde{\mathcal{O}}(\kappa d\sqrt{K})$ &$\widetilde{O}(\kappa d H^{3/2}\sqrt{K})$ \\
	 Lower Bound& $\Omega \left( d\sqrt{K} \right)$  &  $\Omega \left( dH\sqrt{K} \right)$\\
	\hline
\end{tabular}
	\caption{Regret analysis. Upper bounds relative to the PbPO algorithm and information-theoretic lower bounds. Parameters: $d$ = feature dimension, $H$ = sequence horizon, $K$ = number of learning episodes, and $\kappa$ is a coefficient satisfying $\sup_{r_{\rm min} \leq x \leq r_{\rm max}} \vert 1/\sigma'(x) \vert \leq \kappa$.}	\label{tbl:maintho}
	\end{table}

\section{Preference-Based Policy Optimization} \label{sec:offlinekd}
We illustrate the theoretical framework of preference-based policy optimization (PbPO) using a sequence-level reward model and a token-level reward model, respectively.

\begin{algorithm*}[h]
	\caption{PbPO (Theoretical Version)}
	\label{alg:onlinekd}
	\begin{algorithmic}[1]
		\STATE \textbf{Input:} Initial preference set $\mathcal{D}^{\rm pref}_0 = \emptyset$, LLM policy class $\Pi$, initial LLM policy $\pi^0$ (SFT pretrained), input distribution $\mu$
		\FOR{episodes $k = 1$ to $K$}
		\STATE Set the reference policy as the optimized policy in the previous episode: $\pi_{\rm ref}^k \leftarrow \pi^{k-1}$ 
		\STATE	{\bf Step 1: Reward-agnostic exploration}: \\
		\STATE	Choose an enhancer policy $\hat{\pi}^k$ by maximizing the uncertainty measurement:
		\STATE \quad $\hat{\pi}^k = \mathop{\arg\max}_{\pi \in \Pi} \left\Vert \mathbb{E}_{\tau^0 \sim \pi, \tau^1 \sim \pi_{\rm ref}^k} [ \phi(\tau^0) - \phi(\tau^1) ] \right\Vert_{\hat{\Sigma}_k^{-1}}^2$
		\STATE \quad s.t., $\hat{\Sigma}_k = \lambda I + \sum_{s=1}^{k-1} (\phi(\tau^0_s) -  \phi(\tau^1_s))(\phi(\tau^0_s) -  \phi(\tau^1_s))^\top$
		\STATE Sample an input: $x \sim \mu$, trajectories: $\tau^0_k \sim \hat{\pi}^k$, $\tau^1_k \sim \pi_{\rm ref}^k$ 
		\STATE Observe preference label: $o_k \sim \mathbb{P}_{r^\star}(\cdot \mid \tau^0_k, \tau^1_k)$ with an oracle ${r^\star}$ 
		\STATE Add to the preference dataset: $\mathcal{D}^{\rm pref}_k \leftarrow \mathcal{D}^{\rm pref}_{k-1} \cup \{(o_k; \tau^0_k, \tau^1_k)\}$
		\STATE	{\bf Step 2: Reward-aware exploration \& exploitation}: 
		\STATE	Compute the best policy with a constrained min-max optimization problem: 
		\STATE \quad $\pi^k = \mathop{\arg\max}_{\pi \in \Pi} \min_{r \in \mathcal{R}(\mathcal{D}^{\rm pref}_k)} J(\pi, r) - J(\pi_{\rm ref}^k, r)$
		\STATE \quad s.t., $\mathcal{R}(\mathcal{D}^{\rm pref}_k) = \Big\{ r \in \mathcal{G}_r : \sum_{s=1}^{k} \log \mathbb{P}_r (o_s \mid \tau^0_s, \tau^1_s) \geq \max_{r' \in \mathcal{G}_r} \sum_{s=1}^{k} \log \mathbb{P}_{r'} (o_s \mid \tau^0_s, \tau^1_s) - \zeta_k \Big\}$
		\ENDFOR
		\STATE \textbf{Output:} Optimized policy sequence $\{\pi^k\}_{k=1}^K$
	\end{algorithmic}
\end{algorithm*}

\subsection{PbPO with Sequence-Level Reward Model}
In this subsection, we follow \citet{ouyang2022training,rafailov2024direct,xiong2024iterative} and compute the sequence-level reward based on the entire response sequence $y$ given an input $x$. This formulation streamlines the reward model by providing a holistic assessment of the response quality. Note that under the assumption of a sequence-level RM, the MDP problem we previously formulated degenerates into a sentence-wise bandit problem \cite{zhong2025dpo}.  Given a ground-truth sequence-level RM $r^\star$, the trajectory-based preference probability over a binary feedback $o \in \{0,1\}$ for a trajectory pair $(\tau^0, \tau^1)$ is formulated by:
\begin{align*}
	\begin{split}
		\mathbb{P}_{r^\star}(o = 1 \mid \tau^0, \tau^1) = \sigma\left( r^\star(x,y^0) - r^\star(x,y^1) \right).
	\end{split}
\end{align*}

We focus on linear approximation for the sequence-level RM and define the function class using the following assumption:

\begin{assumption}[Linearity \& boundedness of seq-level RM] \label{asp:linearboundseq}
	We assume the RM is linearly parameterized as $r_{\theta}(x,y) = \langle \theta, \phi(x,y) \rangle$, where $\phi: \tau \mapsto \phi(x,y) \in \mathbb{R}^d$ is a fixed sequence-level feature extractor. For regularization, the features and parameters are bounded such that $\sup_{(x,y)} \Vert \phi(x,y) \Vert_2 \leq 1$ and $\Vert \theta \Vert_2 \leq B$ for some $B > 0$.
\end{assumption}

We consider a function approximation approach for estimating the ground-truth RM $r^\star$. Specifically, we introduce a linear function class $\mathcal{G}_r^{\rm seq}$ to approximate $r^\star$:
\begin{align} \label{eq:rmclassseq}
	\begin{split}
	\mathcal{G}_r^{\rm seq} := \Bigl\{ & r(x,y) = \theta^\top \phi(x,y): \\
	& \quad \|\theta\|_2 \le B, \|\phi(x,y)\|_2 \le 1 \Bigr\}.
	\end{split}
\end{align}

We assume that the RM class satisfies realizability:
\begin{assumption}[Realizability of sequence-level RM] \label{asp:reaseq}
	We assume the reward class is realizable, i.e., the ground-truth sequence-level RM lies in the function class: $r^\star \in \mathcal{G}_r^{\rm seq}$, i.e., it satisfies that $r^\star(x,y) = \langle \theta^\star, \phi(x,y) \rangle$ for some $\Vert \theta^\star \Vert \leq B$.
\end{assumption}

\subsubsection{Algorithm} 
We introduce the PbPO algorithm in Algorithm \ref{alg:onlinekd}. 
During training episodes $k \in \{1, 2, \ldots, K\}$, preference feedback can be adaptively collected. This enables the LLM policy to continuously refine itself based on up-to-date information, thereby achieving bootstrapping performance. At each episode, the reference policy $\pi_{\rm ref}^k$ is defined as the optimized policy from the previous round, then the algorithm mainly consists of two steps as follows. 

{\bf Step 1}: Reward-agnostic exploration by collecting trajectories with an enhancer policy (Lines 4--10). To learn the ground-truth RM, we collect exploratory trajectories that cover the space spanned by $\phi(\cdot)$ before collecting any human feedback. To achieve this we identify a set of explorative enhancer policy that are not covered by existing preference data from the previous episodes. We measure the extent to which the trajectory generated by $(\hat{\pi}^k, \pi^k_{\rm ref})$ can be covered by computing the norm of $\mathbb{E}_{\tau^0 \sim \hat{\pi}_k, \tau^1 \sim \pi_{\rm ref}^k} [ \phi(\tau^0) - \phi(\tau^1) ]$ on the metric induced by the inverse empirical covariance matrix $\hat{\Sigma}_k$ at the $k$-th episode. This strategy encourages querying trajectory pairs $(\tau_k^0, \tau_k^1)$ that highlight uncertain regions of the RM. The obtained preference-labeled pair $(o_k; \tau_k^0, \tau_k^1)$ is then added to the cumulative dataset $\mathcal{D}_k^{\rm pref}$. The process can be viewed as reward-agnostic exploration.

{\bf Step 2}:  Reward-aware exploration and exploitation by solving a min–max optimization problem (Lines 11--14).
Given the collected preference dataset $\mathcal{D}_{k}^{\rm pref}$, the objective presented in Lines 13--14 constitutes a min-max optimization problem between the policy and an uncertain RM. The outer objective is to optimize a target LLM policy $\pi^k$ by maximizing its performance gap with a reference policy $\pi_k^{\rm ref}$ ($=\pi^{k-1}$), while the inner minimization identifies the least favorable reward model $r \in \mathcal{R}(\mathcal{D}^{\rm pref}_{k})$ that minimizes the performance gap. This ensures conservative yet guaranteed improvement relative to the reference policy, while accounting for the inherent uncertainty of the reward inference process. Specifically, the reward confidence set $\mathcal{R}(\mathcal{D}^{\rm pref}_{k}) \subseteq \mathcal{G}_r$ (defined in Line 14) is constructed via the maximum likelihood estimation (MLE) on the preference dataset $\mathcal{D}^{\rm pref}_{k}$ with $\zeta_k \geq 0$ being a slack parameter controlling the confidence radius, thereby encouraging reward-aware exploration for a distributionally robust formulation. Since the RM can be constrained near the best empirical RM, the optimization of policy gains exploitation from the previous preference data.

\subsubsection{Learning Guarantee} 
We provide a theoretical guarantee for Algorithm~\ref{alg:onlinekd} by establishing a cumulative regret bound of Eq. (\ref{eq:regret}). 

\begin{theorem}[Regret bound with sequence-level RM] \label{thm:regretseq}
	For any $\delta \in (0,1]$, let $\zeta_k = \mathcal{O}(d\log(Bk/\delta))$ for any $k \in \{1,2,\ldots,K\}$ with a maximum episode number of $K$, then under Assumptions \ref{asp:linearboundseq} \& \ref{asp:reaseq}, setting $\gamma = \sqrt{\log\left( 1 + 4K/(c_2d^2 \log(K/\delta)) \right)}$ we have with probability at least $1-3\delta$:
	\begin{align}
		{\rm Regret}(K) \leq c_1 \gamma B \kappa d\sqrt{K}\log(BK/\delta),
	\end{align}
	where $c_1, c_2 > 0$ denote some universal constants.
\end{theorem}

\begin{remark}
	Theorem~\ref{thm:regretseq} establishes that Algorithm~\ref{alg:onlinekd} achieves a regret bound of order $\widetilde{\mathcal{O}}(\kappa d\sqrt{K})$ with high probability, where $K$ is the number of episodes and $d$ is the feature dimension. If neglecting $\kappa$, this result is analogous to the regret bounds in standard linear bandit settings, such as those achieved by the LinUCB algorithm for exploration \cite{dani2008stochastic}.
\end{remark}

\begin{corollary}[Sample complexity with sequence-level RM] \label{onlinecomplexseq}
	Under the conditions of Theorem~\ref{thm:regretseq}, if the number of preference samples satisfies $K \;\geq\; \widetilde{\mathcal{O}}\!\left(\kappa^2 d^2/\epsilon^2\right)$, then Algorithm~\ref{alg:onlinekd} achieves $\epsilon$-approximate convergence with high probability; that is, there exists some $k_0 \in \{1,2,\ldots,K\}$ such that $J(\pi^\star, r^\star) - J(\pi^{k_0}, r^\star) \leq \epsilon$ with high probability. Since the algorithm generates exactly one preference sample per episode, the sample complexity is equivalent to the number of episodes $K$.
\end{corollary}

To show that the regret bound is nearly optimal, we use the follow information-theoretic lower bound.

\begin{theorem}[Regret lower bound with sequence-level RM] \label{thm:lowerseq}
	Under Assumptions \ref{asp:linearboundseq} \& \ref{asp:reaseq}, there exists a reward model $r^\star \in \mathcal{G}_r^{\rm seq}$ such that for any algorithm, the expected pseudo-regret over $K \geq \Omega(d^2/B^2)$ episodes is lower bounded as:
	\begin{align}
		\mathbb{E}_{\theta^\star} \left[{\rm Regret}(K)\right] \geq \Omega \left( d\sqrt{K} \right).
	\end{align}
\end{theorem}

\subsection{PbPO with Token-Level Reward Model}
In this subsection, we focus on token-level RM, as discussed in recent RLHF advances based on token-level MDPs \cite{zeng2024token,cen2025value,zhong2025dpo}.

Given the ground-truth token-level RM $r^\star=\{r^\star_h\}_{h=1}^H$, the trajectory-based preference probability for trajectory pairs $\tau^0 = \{s_1,a^0_1, s^0_2, a^0_2, \ldots, s^0_H, a^0_H \}$ and $\tau^1 = \{s_1, a^1_1, s^1_2, a^1_2, \ldots, s^1_H, a^1_H \}$ over a binary feedback $o \in \{0,1\}$ is represented as:
\begin{align*}
	\mathbb{P}_{r^\star}(o = 1 | \tau^0, \tau^1) = \sigma \left(\sum_{h=1}^H \!\! r^\star_{h}(s^0_h, a^0_h) - \sum_{h=1}^H \!\! r^\star_{h}(s^1_h, a^1_h)\!\! \right)\!\!.
\end{align*}

We focus on linear approximation for the token-level RM with the following assumption:
\begin{assumption}[Linearity \& boundedness of token-level RM] \label{asp:linearboundtoken}
	We assume that the token-level RM is parameterized by ${\theta}=(\theta_1,\theta_2,\ldots,\theta_H) \in \mathbb{R}^{dH}$ and defined as $r_{{\theta}}: \tau \mapsto \sum_{h=1}^{H} \theta_h^\top \phi(s_h,a_h) \in \mathbb{R}$, where $H$ denotes the sequence length. 
	For each $h \in [H]$, we assume the RM is represented as $r_{\theta_h}(s_h, a_h) = \langle \theta_h, \phi(s_h, a_h) \rangle$. For regularization, the step-wise parameters $\theta_h \in \mathbb{R}^d$ are bounded as $\Vert \theta_h \Vert_2 \leq B$, and the features satisfy $\phi(s_h, a_h) \in \mathbb{R}^d$ with $\sup_{s_h,a_h} \Vert \phi(s_h, a_h) \Vert_2 \leq 1$.
\end{assumption}

We consider a function approximation approach for estimating the ground-truth RM $r^\star=\{r^\star_h\}_{h=1}^H$. Specifically, we introduce a linear function class $\mathcal{G}_r^{\rm tok}$ to approximate $r^\star$:
\begin{align} \label{eq:rmclasstoken}
	\begin{split}
	\mathcal{G}_{r}^{\rm tok} := \Bigl\{ \{r_h\}_{h=1}^H \,\big|\, 
	r_h(s,a) = \theta_h^\top \phi(s,a): \\
	\|\theta_h\|_2 \le B,\; \|\phi(s,a)\|_2 \le 1 \Bigr\}.
	\end{split}
\end{align}

We define the realizability of RM with the following assumption:
\begin{assumption}[Realizability of token-level RM] \label{asp:reaonlinetoken}
	We assume the reward class is realizable, i.e., the ground-truth RM lies in the function class: $r^\star \in \mathcal{G}_r^{\rm tok}$, i.e., for each $h 
	\in [H]$, it satisfies that $r^\star_h(s_h, a_h) = \langle \theta^\star_h, \phi(s_h, a_h) \rangle$ for some $\Vert \theta^\star_h \Vert \leq B$.
\end{assumption}

\subsubsection{Algorithm} 
The theoretical learning procedure is summarized in Algorithm \ref{alg:onlinekd}, which follows the same structure as the sequence-level RM setting, with the only differences being the definition of the reward model and the feature projection: $\phi(\tau) = [\phi(s_1,a_1), \phi(s_2, a_2), \ldots, \phi(s_H, a_H)]$.

\subsubsection{Learning Guarantee}
We present the regret bound and sample complexity of PbPO under the token-level RM setting as follows.
\begin{theorem}[Regret bound with token-level RM] \label{thm:regretupperaction}
	For any $\delta \in (0,1]$, let $\zeta_k = \mathcal{O}(dH\log(Bk\sqrt{H}/\delta))$ for any $k \in \{1,2,\ldots,K\}$ with a maximum episode number of $K$, then under Assumptions \ref{asp:linearboundtoken} \& \ref{asp:reaonlinetoken}, setting $\gamma = \sqrt{\log\left( 1 + 4K/(c_2Hd^2 \log(K/\delta)) \right)}$, we have with probability at least $1-3\delta$:
	\begin{align}
		{\rm Regret}(K) \leq c_1 \gamma B \kappa d  H^{3/2} \sqrt{K} \log(B\sqrt{H}K/\delta),
	\end{align}
	where $c_1, c_2 > 0$ denote some universal constants.
\end{theorem}

\begin{remark}
	Theorem \ref{thm:regretupperaction} establishes a high probabiliy regret bound of order $\widetilde{O}(\kappa d H^{3/2}\sqrt{K})$ for PbPO with token-level reward model.
\end{remark}

Based on the upper bound of regret, we can derive the following corollary on the sample complexity for the PbPO algorithm with token-level RM:
\begin{corollary}[Sample complexity with token-level RM]
	To achieve an $\epsilon$-approximate convergence, i.e., there exists some $k_0 \in \{1,2,\ldots,K\}$ such that $J(\pi^\star, r^\star) - J(\pi^{k_0}, r^\star) \le \epsilon$,
	it suffices to collect $K = \widetilde{\mathcal{O}}\!\left(\kappa^2 d^2 H^3/\epsilon^2 \right)$ preference samples. Since our algorithm generates exactly one trajectory-based preference feedback per episode, the total number of episodes $K$ coincides with the sample complexity. 
\end{corollary}

\begin{theorem}[Regret lower bound with token-level RM] \label{thm:lowertok}
	Under Assumptions \ref{asp:linearboundtoken} \& \ref{asp:reaonlinetoken}, there exists a reward model $r^\star \in \mathcal{G}_r^{\rm tok}$ such that for any algorithm, the expected pseudo-regret over $K \geq \Omega(d^2/B^2)$ episodes is lower bounded as follows:
	\begin{align}
		\mathbb{E}_{\theta^\star} \left[{\rm Regret}(K)\right] \geq \Omega \left( dH\sqrt{K} \right).
	\end{align}
\end{theorem}

This theorem demonstrates that our regret bound is nearly optimal, with only a small gap of $\widetilde{\mathcal{O}}(\kappa \sqrt{H})$.

\subsection{Approximate Policy Optimization for PbPO}
In this section, we describe how to implement the theoretical PbPO algorithms in practice.

\begin{algorithm}[t]
	\caption{PbPO (Practical Version)}
	\label{alg:pratical}
	\begin{algorithmic}[1]
		\STATE \textbf{Input:} Input dataset $\mathcal{D}_{\rm in}$, number of episodes $K$, batch size $\mathcal{B}$, outer step size $T_{\rm out}$, inner step size $T_{\rm in}$. Initialize $\hat{\Sigma}_0 = \lambda I$, $\mathcal{D}^{\rm pref}_0 = \emptyset$, $\pi^0 = \pi_{\rm SFT}$
		\FOR{episodes $k = 1$ to $\lfloor K/\mathcal{B} \rfloor$}
		\STATE Set the reference policy as the optimized policy in the previous episode: $\pi_{\rm ref}^k \leftarrow \pi^{k-1}$ 
		\STATE	{\bf Step 1: Reward-agnostic exploration}: \\
		\STATE Update $\hat{\Sigma}_k \leftarrow \hat{\Sigma}_{k-1} + \sum_{j=1}^\mathcal{B} (\phi(\tau^0_{k-1,j}) -  \phi(\tau^1_{k-1,j}))(\phi(\tau^0_{k-1,j}) -  \phi(\tau^1_{k-1,j}))^\top$
		\STATE	Maximize enhancer policy:
		\STATE $\hat{\pi}^k \leftarrow \mathop{\arg\max}\limits_{\pi \in \Pi} \left\Vert \mathbb{E}_{\tau^0 \sim \pi, \tau^1 \sim \pi_{\rm ref}^k} [ \phi(\tau^0) - \phi(\tau^1) ] \right\Vert_{\hat{\Sigma}_k^{-1}}^2$
        \FOR{$j = 1$ to $\mathcal{B}$}
        \STATE Sample input $x_j \in \mathcal{D}_{\rm in}$, trajectories $\tau^0_{k,j} \sim \hat{\pi}^k$, $\tau^1_{k,j} \sim \pi_{\rm ref}^k$, observe preference label $o_{k,j} \sim \mathbb{P}_{r^\star}(\cdot \mid \tau^0_{k,j}, \tau^1_{k,j})$
        \STATE Add $\mathcal{D}^{\rm pref}_k \leftarrow \mathcal{D}^{\rm pref}_{k-1} \cup \{(o_{k,j}; \tau^0_{k,j}, \tau^1_{k,j})\}$
        \ENDFOR 
		\STATE	{\bf Step 2: Reward-aware exploration \& exploitation}: 
	    \FOR{outer steps $t=1$ to $T_{\rm out}$}
	    \STATE $\pi^k \leftarrow \operatorname{SGA}\big(J(\pi^k, \hat{r})\big)$ $\hfill \rhd$ Eq. (\ref{eq:stackelbergobjout})
	    \FOR{inner steps $t'=1$ to $T_{\rm in}$}
	    \STATE $\hat{r}  \leftarrow \operatorname{SGD} \big( J(\pi, \hat{r}) - J(\pi_{\rm ref}^k, \hat{r})  -  \beta \sum_{n=1}^{|\mathcal{D}^{\rm pref}_k|} \log \mathbb{P}_{\hat{r}}(o_n \mid \tau^0_n, \tau^1_n) \big)$ $\hfill \rhd$ Eq. (\ref{eq:stackelbergobjin})
	    \ENDFOR    
	    \ENDFOR
	    \ENDFOR
		\STATE \textbf{Output:} Optimized policy sequence $\{\pi^k\}_{k=1}^K$
 	\end{algorithmic}
\end{algorithm}

\subsubsection{Stackelberg Game Formulation of the Min-Max Objective}  
The original objective in Algorithm~\ref{alg:onlinekd} (Lines 13--14) defines a constrained min-max optimization problem over the LLM policy and the RM. However, solving this problem is generally intractable when employing flexible function approximators such as neural networks. To address this issue, we reformulate the objective as a two-player Stackelberg game \cite{von2010market} between the LLM policy (leader) and the RM (follower).  

To circumvent the difficulty of directly optimizing under the RM’s confidence-set constraint, we apply a Lagrangian relaxation. Specifically, we introduce a Lagrange multiplier $\beta \ge 0$ and convert the constrained min-max problem into an unconstrained bi-level optimization problem:
\begin{align}
	\begin{split} \label{eq:stackelbergobjout}
		\hat{\pi} &\in \mathop{\arg\max}_{\pi \in \Pi} \; J(\pi, r^{\pi}) - J(\pi_{\rm ref}, r^{\pi}), \\
	\end{split}
\end{align}
such that
\begin{align}
	\begin{split} \label{eq:stackelbergobjin}
		 r^{\pi}  &\in \mathop{\arg\min}_{{r} \in \mathcal{G}_r} 
		\Big\{ J(\pi, r) - J(\pi_{\rm ref}, r)  \\
		&\qquad\qquad\quad - \beta \sum_{n=1}^{|\mathcal{D}^{\rm pref}|} 
		\log \mathbb{P}_{r}(o_n \mid \tau^0_n, \tau^1_n) \Big\}.
	\end{split}
\end{align}

Here, the RM $r$ is optimized to maximize the likelihood of the observed preferences via maximum likelihood estimation (MLE) based on the preference dataset. This dataset, denoted by $\mathcal{D}^{\rm pref}_k$, is iteratively updated at each online training episode $k \in \{1,2,\ldots,K\}$ in Algorithm~\ref{alg:onlinekd}.  

\subsubsection{Gradient-based Policy Optimization}  
Following the two-player game formulation, we adopt a gradient-based adversarial training procedure for policy optimization, as illustrated in Algorithm \ref{alg:pratical}. At the beginning of each online episode $k$, after optimizing the enhancer policy, input samples are sampled from the input dataset. Preference data are collected by sampling trajectories from both the the enhancer policy and the reference policy, which are then aggregated with previous preference data (Lines 8--11).

The two-player game is solved via gradient-based adversarial training for policy optimization (Lines 13--18). Specifically, the main LLM policy is updated via stochastic gradient ascent (SGA) to maximize the objective in Eq. (\ref{eq:stackelbergobjout}) (Line 14), while the reward model (RM) is updated via stochastic gradient descent (SGD) to minimize the objective in Eq. (\ref{eq:stackelbergobjin}) (Line 16).

\section{Experiments}
We evaluate the PbPO approach on five benchmarks.

\begin{table*}[t]
	\setlength{\tabcolsep}{1.2pt}
	\centering
	\begin{tabular}{l|cccccc|cccccc}
		\hline
		\multirow{2}{*}{\bf Method} & \multicolumn{6}{c|}{\bf LLaMA2-7B} 
		& \multicolumn{6}{c}{\bf Qwen2-7B} \\
		& BBH & AGIEval & ARC-C & MMLU & GSM8K & {\bf Avg.} 
		& BBH & AGIEval & ARC-C & MMLU & GSM8K & {\bf Avg.}  \\
		\hline
		
		SFT 
		& 43.6$_{\pm .3}$ & 34.3$_{\pm .3}$ & 66.7$_{\pm .5}$ & 48.3$_{\pm .4}$ & 49.3$_{\pm .2}$ & 48.4  
		& 61.4$_{\pm .2}$ & 50.0$_{\pm .2}$ & 61.5$_{\pm .4}$ & 69.5$_{\pm .3}$ & 75.5$_{\pm .3}$ & 63.6  \\
		DPO 
		& 45.2$_{\pm .4}$ & 35.3$_{\pm .3}$ & 67.6$_{\pm .5}$ & 49.5$_{\pm .4}$ & 51.0$_{\pm .4}$ & 49.7
		& 65.3$_{\pm .3}$ & 52.7$_{\pm .4}$ & 63.2$_{\pm .5}$ & 70.5$_{\pm .3}$ & 77.8$_{\pm .3}$ & 65.9   \\
		
		PPO 
		&44.7$_{\pm .5}$&33.7$_{\pm .2}$&68.0$_{\pm .4}$&50.2$_{\pm .4}$&49.5$_{\pm .2}$& 49.2
		&64.7$_{\pm .5}$&54.9$_{\pm .4}$&65.3$_{\pm .3}$&72.7$_{\pm .3}$&78.2$_{\pm .5}$&67.2   \\ 
		Online DPO (300 eps)
		& 45.6$_{\pm .5}$ & 35.6$_{\pm .3}$ & 68.7$_{\pm .4}$ & 52.3$_{\pm .4}$ & 51.7$_{\pm .4}$ & 50.8 
		& 65.6$_{\pm .3}$ & 55.7$_{\pm .2}$ & 68.2$_{\pm .5}$ & 70.2$_{\pm .4}$ & 76.5$_{\pm .4}$ &67.2  \\
		
		Online PPO (300 eps)
		&46.2$_{\pm .5}$&35.5$_{\pm .4}$&66.7$_{\pm .6}$&50.0$_{\pm .7}$&51.0$_{\pm .6}$& 49.9
		&66.2$_{\pm .6}$&54.2$_{\pm .4}$&65.2$_{\pm .6}$&73.2$_{\pm .3}$&79.2$_{\pm .6}$&67.6   \\
		
		Best-of-$N$ Distill (300 eps)
		& 46.8$_{\pm .3}$ & 33.8$_{\pm .4}$ & 67.5$_{\pm .3}$& 49.2$_{\pm .2}$ & 50.7$_{\pm .4}$ & 49.6
		& \textbf{68.2}$_{\pm .4}$ & 53.2$_{\pm .5}$ & 65.4$_{\pm .3}$& 72.7$_{\pm .2}$ & 78.2$_{\pm .4}$ &67.5   \\
		\hline
		PbPO w/ seq RM (100 eps)
		&49.4$_{\pm .4}$&40.5$_{\pm .3}$&68.3$_{\pm .5}$&53.2$_{\pm .3}$&52.1$_{\pm .6}$&52.7
		&65.7$_{\pm .3}$&57.8$_{\pm .4}$&66.7$_{\pm .4}$&73.4$_{\pm .5}$&81.4$_{\pm .4}$& 69.0  \\
		
		PbPO w/ seq RM (200 eps)
		&52.4$_{\pm .3}$&\textbf{44.0}$_{\pm .2}$&{71.6}$_{\pm .4}$&{55.4}$_{\pm .5}$&{55.2}$_{\pm .4}$&55.7
		&66.3$_{\pm .4}$&\underline{59.2}$_{\pm .5}$&{68.7}$_{\pm .5}$&{75.0}$_{\pm .3}$&{82.5}$_{\pm .3}$&{70.3}   \\
		
		PbPO w/ seq RM (300 eps)
		&\underline{53.9}$_{\pm .6}$&\underline{43.7}$_{\pm .3}$&\underline{72.9}$_{\pm .5}$&\textbf{57.1}$_{\pm .6}$&\underline{58.1}$_{\pm .4}$&\underline{57.1} 
		&{67.8}$_{\pm .4}$&{58.2}$_{\pm .4}$&\underline{70.5}$_{\pm .4}$&\textbf{77.2}$_{\pm .4}$&{85.2}$_{\pm .3}$&\underline{71.8}  \\
		
		PbPO w/ tok RM (100 eps)
		&49.0$_{\pm .5}$&39.2$_{\pm .5}$&68.8$_{\pm .4}$&52.0$_{\pm .4}$&54.5$_{\pm .4}$&52.7
		&66.3$_{\pm .4}$&57.2$_{\pm .6}$&67.5$_{\pm .3}$&71.5$_{\pm .3}$&83.6$_{\pm .3}$& 69.2  \\
		
		PbPO w/ tok RM (200 eps)
		&52.7$_{\pm .4}$&{41.7}$_{\pm .3}$&{72.0}$_{\pm .3}$&{54.8}$_{\pm .5}$&{57.2}$_{\pm .5}$&55.7
		&67.2$_{\pm .3}$&{58.1}$_{\pm .4}$&{69.5}$_{\pm .4}$&{74.6}$_{\pm .2}$&\underline{85.8}$_{\pm .2}$&{71.0}   \\
		
		PbPO w/ tok RM (300 eps)
		&\textbf{54.3}$_{\pm .5}$&{43.5}$_{\pm .4}$&\textbf{73.5}$_{\pm .3}$&\underline{56.3}$_{\pm .4}$&\textbf{60.2}$_{\pm .4}$&\textbf{57.6} 
		&\underline{68.0}$_{\pm .3}$&\textbf{59.5}$_{\pm .4}$&\textbf{71.0}$_{\pm .3}$&\underline{76.4}$_{\pm .6}$&\textbf{87.2}$_{\pm .4}$&\textbf{72.4}  \\
		
		\hline
	\end{tabular}
	\caption{Main results on LLaMA2-7B and Qwen2-7B backbones. 
		We format {\bf the best} and \underline{the second best} results.}	\label{tbl:main}
\end{table*}

\subsection{Experimental Setup}
\noindent\textbf{Benchmarks.}  
We evaluate on a suite of benchmarks including the complex reasoning dataset BBH \cite{suzgun2023challenging}, knowledge-based benchmarks AGIEval \cite{zhong2024agieval}, ARC-C \cite{clark2018think}, and MMLU \cite{hendryckstest2021}, as well as the math reasoning benchmark GSM8K \cite{cobbe2021training}. All reported accuracy results are averaged over three random seeds $\{10, 20, 30\}$ using zero-shot decoding.

\noindent\textbf{Preference data collection.}  
In each online learning episode, we randomly sample a batch of prompts from the training set to generate preference pairs using outputs from the reference policy and the enhancer policy. The corresponding preference labels are derived from GPT-4 feedback. To construct the offline preference dataset, we utilize models from the LLaMA2 \cite{touvron2023llama} and Qwen2 \cite{qwen2} families, including LLaMA2-7B/13B/70B-Chat and Qwen2-1.5B/7B/72B to generate candidate response pairs. 


\noindent\textbf{Hyperparameters.}  
For both the initial policy and the enhancer policy, we employ a pretrained LLaMA2-7B \cite{touvron2023llama} or Qwen2-7B \cite{qwen2} model, fine-tuned on the supervised training dataset. The reward model is constructed by appending a linear output layer to a frozen LLaMA2-7B or Qwen2-7B backbone.
For the main experiments, we conduct a total of 300 episodes of online PbPO. In each episode, preference data are sampled with a batch size of 32. 

\noindent\textbf{Baselines.}  
We compare our approach against several strong offline and online preference optimization baselines:

\begin{itemize}
	\item \textbf{DPO:} Performing Direct Preference Optimization (DPO) \cite{rafailov2024direct} using the offline preference dataset.
	
	\item \textbf {PPO:} Training a reward model using the offline preference dataset, followed by PPO fine-tuning \cite{ouyang2022training}.
	
	\item \textbf{Online DPO:} Similar to Self-Rewarding LM \cite{yuan2024self}, where the main LLM policy generates new preference pairs for iterative fine-tuning via DPO.
	
	\item \textbf{Online PPO:} Following \citet{xiong2024iterative}, iterative RLHF training where new preference pairs are generated and added to the dataset for reward model learning at each episode.
	
	\item \textbf{Best-of-$N$ Distill:} Following \cite{sessabond}, we perform iterative self-distillation. In each episode, $N = 10$ candidate completions are generated using the teacher policy via nucleus sampling (top-p = 0.9, top-k = 40, temperature = 0.8). The response with the highest reward score is used as the training signal for knowledge distillation with KL divergence.
\end{itemize}

\subsection{Main Results}  
Table~\ref{tbl:main} presents the main results. Compared to the SFT baseline, all preference-based optimization methods yield noticeable improvements, confirming the effectiveness of leveraging preference data for LLM fine-tuning. However, existing online approaches—including Online DPO, Online PPO, and Best-of-$N$ Distillation—only marginally outperform the offline RLHF baselines.
In contrast, our proposed method consistently surpasses both offline and online baselines across all five evaluation datasets, demonstrating the superiority of our preference-based policy optimization (PbPO) framework in enhancing policy optimization.
We conduct an ablation study on the number of online PbPO episodes. The results indicate that online PbPO reliably outperforms its offline variant, underscoring the advantage of iterative preference alignment. Performance continues to improve with more episodes, highlighting the effectiveness of bootstrapping LLMs through repeated preference-based refinement. Additionally, we observe that PbPO based on token-level RM achieves superior final performance compared to its sequence-level counterpart, particularly on datasets that require multi-step reasoning such as BBH, ARC-C, and GSM8K. This shows that token-level RM has an advantage for mastering complex, long-horizon tasks.

\subsection{Experimental Analysis}
\begin{figure*}[t]
	\centering
	\hspace{0.05in}
	\subfigure[Accuracy on AGIEval.]{
		\begin{minipage}[h]{0.31\textwidth}
			\centering
			\includegraphics[width=5.7cm]{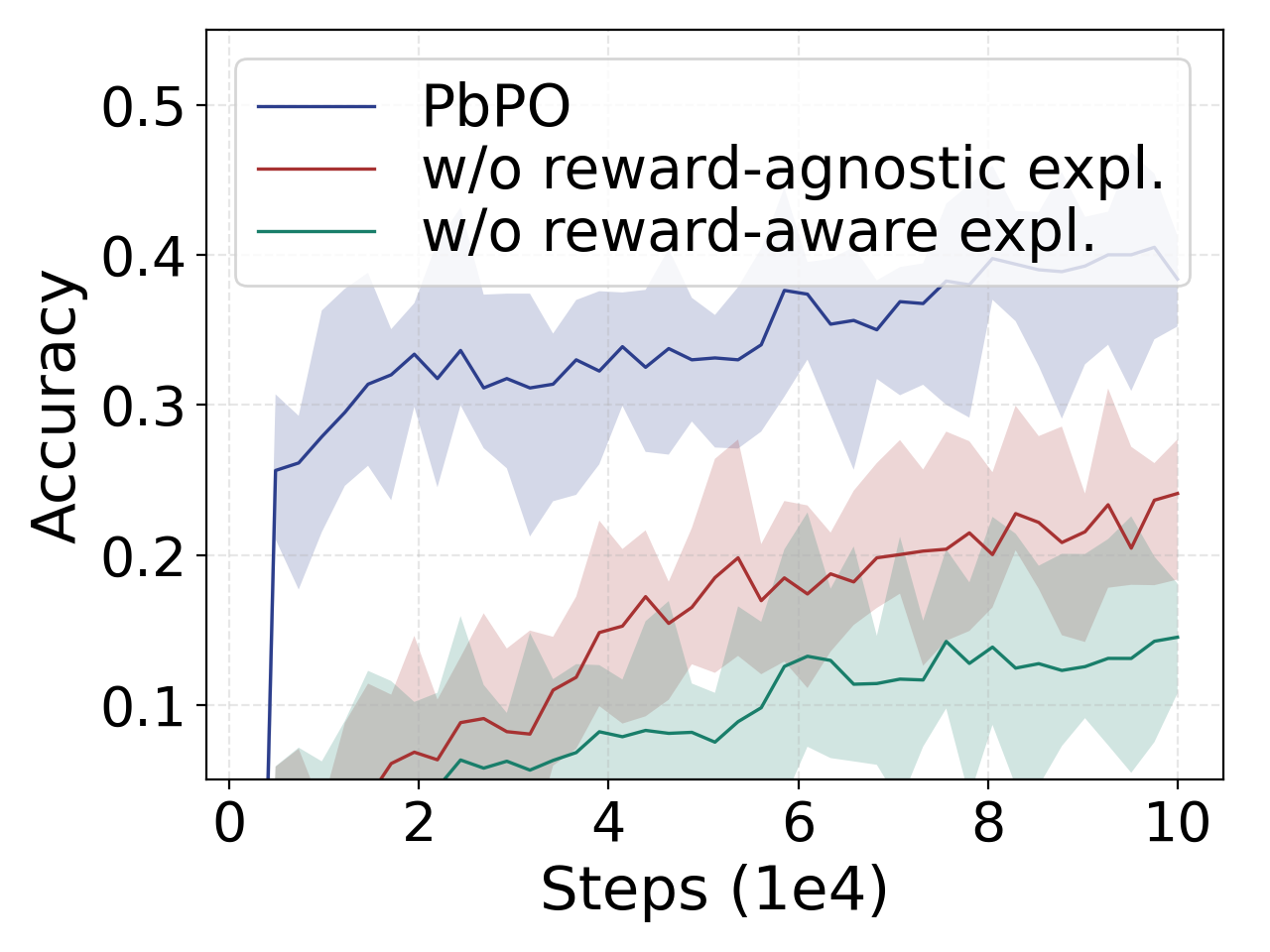}
	\end{minipage}} 	
	\hspace{0.05in}
	\subfigure[Accuracy on ARC-C.]{
		\begin{minipage}[h]{0.31\textwidth}
			\centering
			\includegraphics[width=5.7cm]{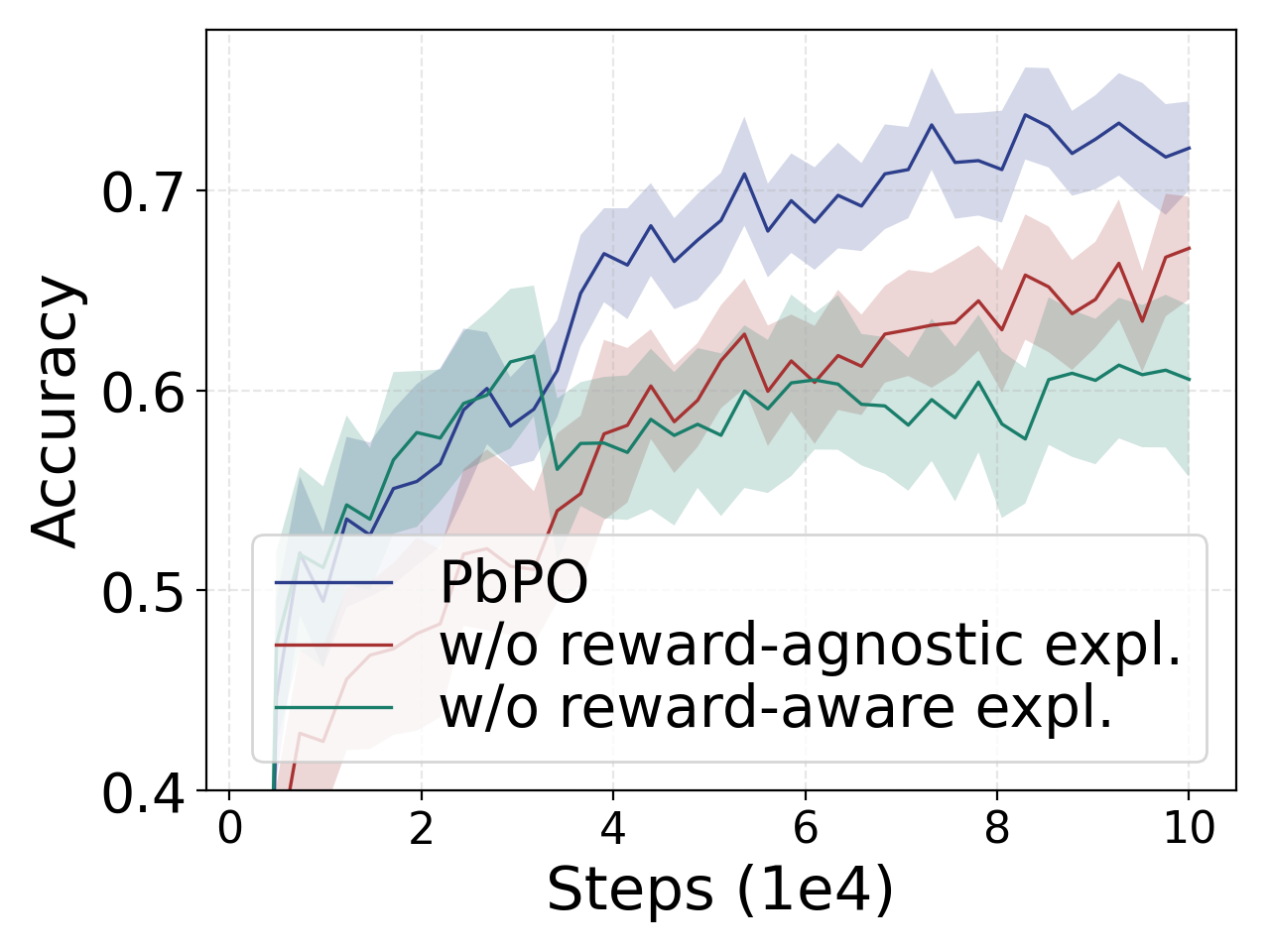}
	\end{minipage}} 
	\hspace{0.05in}
	\subfigure[Accuracy on MMLU.]{
		\begin{minipage}[h]{0.31\textwidth}
			\centering
			\includegraphics[width=5.7cm]{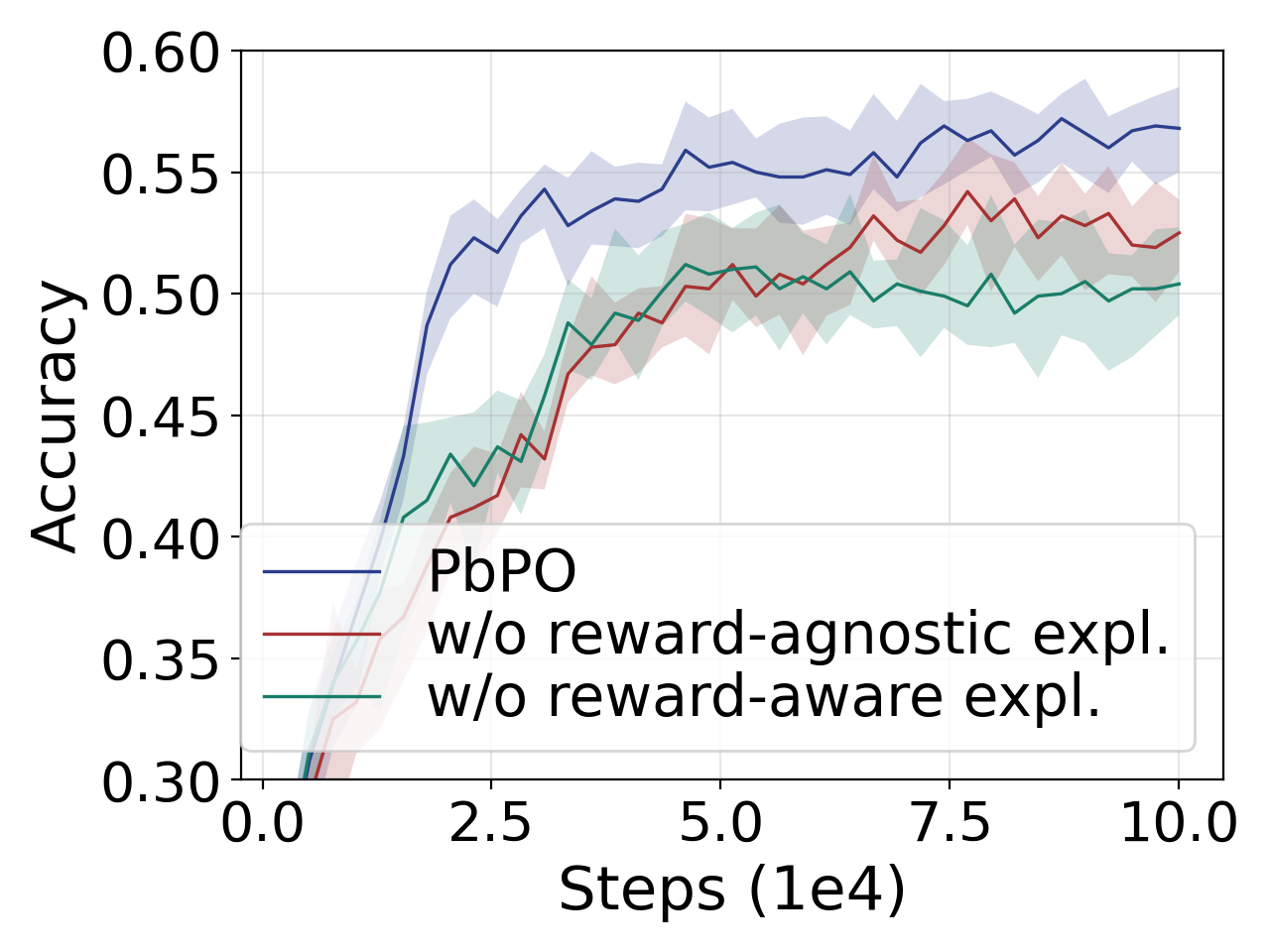}
	\end{minipage}} 
	\hspace{0.05in}
	\subfigure[Accuracy on GSM8K.]{
		\begin{minipage}[h]{0.31\textwidth}
			\centering
			\includegraphics[width=5.7cm]{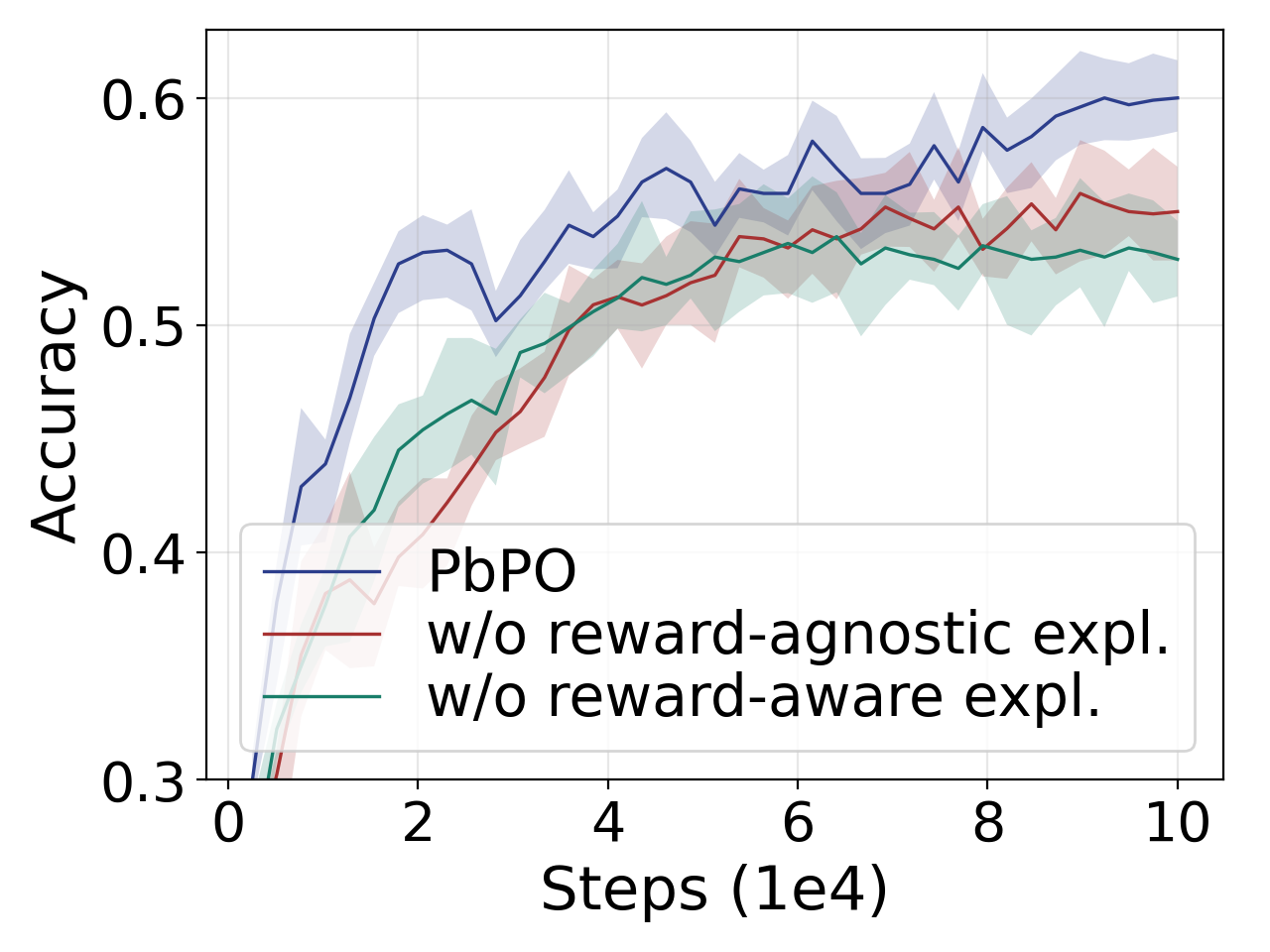}
	\end{minipage}} 
	\hspace{0.05in}
	\subfigure[Sensitivity to conservatism $\beta$.]{
		\begin{minipage}[h]{0.31\textwidth}
			\centering
			\includegraphics[width=5.7cm]{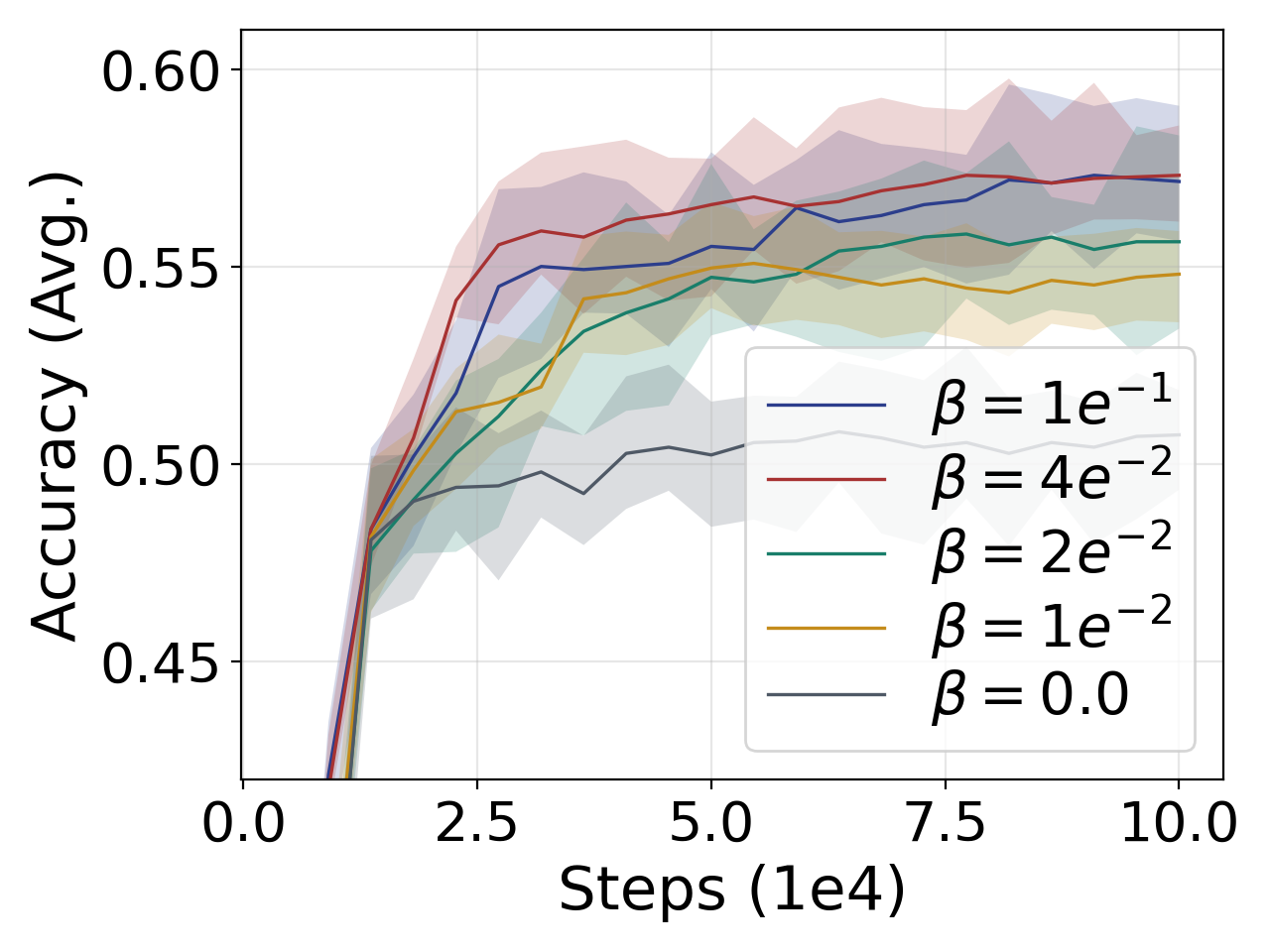}
	\end{minipage}} 
	\hspace{0.05in}
	\subfigure[Effect of reward model size.]{
		\begin{minipage}[h]{0.31\textwidth}
			\centering
			\includegraphics[width=5.7cm]{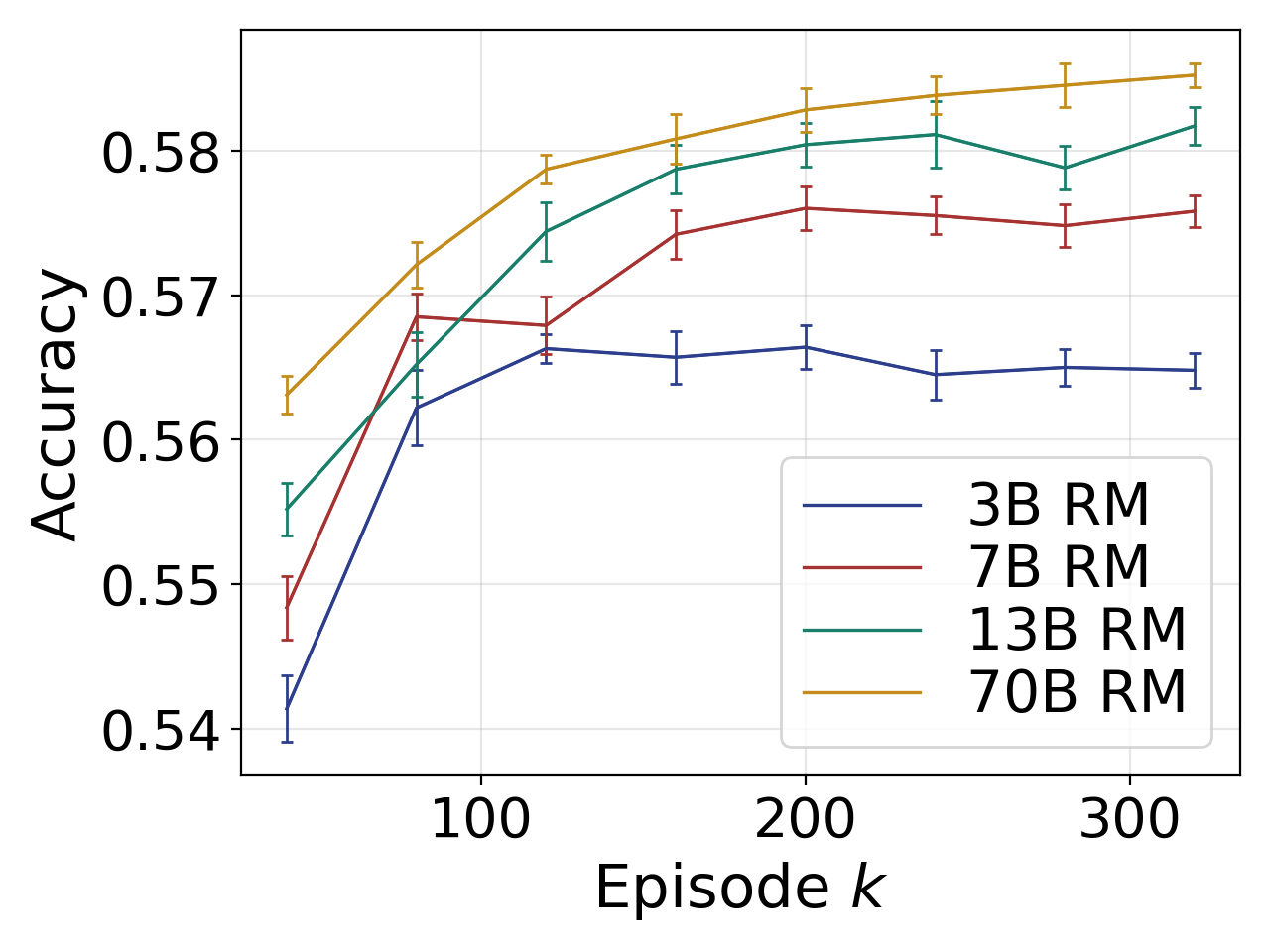}
	\end{minipage}} 
	\caption{Experimental analysis based on LLaMA2-7B backbone.} 
	\label{fig:dist1}
\end{figure*}

\noindent\textbf{Ablation study on exploration strategies.}
We evaluate PbPO against two ablations:  
(i) \textbf{PbPO w/o reward-agnostic explor.}, which removes the optimized enhancer and instead uses only the reference policy to generate trajectory pairs for preference data; and  
(ii) PbPO w/o reward-aware exploration, which follows a pipeline approach: it first trains the reward model using the online preference data and then performs policy optimization in each episode (similar to \citet{xiong2024iterative}). 
Figure~\ref{fig:dist1}(a--d) shows that our full method converges to a higher plateau across all tasks.  
The ablation without reward-agnostic exploration achieves lower final performance, while the one without reward-aware exploration performs significantly worse, confirming that both exploration components are critical.

\noindent\textbf{Sensitivity of the conservatism regularizer $\beta$.}
Figure~\ref{fig:dist1}(e) shows the effect of $\beta$, which balances adversarial and preference losses in Eq.~(\ref{eq:stackelbergobjin}).  
Any $\beta > 0$ leads to steady improvement and higher stability compared to $\beta = 0$, where the RM fails to learn effectively.  
With $\beta \in \{0, 0.01, 0.02, 0.04, 0.1\}$, larger values ($0.04$, $0.1$) achieve similar top performance, indicating robustness to $\beta$ selection without fine-tuning.

\noindent\textbf{Effect of RM size.}
We test RM sizes from OpenLLaMA-3B to LLaMA2-70B (Figure~\ref{fig:dist1}(f)).  
All sizes benefit from more episodes, confirming the advantage of online learning.  
Smaller RMs (e.g., 3B) converge faster but plateau lower; larger RMs achieve better final performance, suggesting greater RM capacity improves preference alignment.

\section{Related Work}
\noindent\textbf{RLHF.}  
Reinforcement learning from human feedback (RLHF) has become a pivotal approach for fine-tuning large language models (LLMs) to produce text better aligned with human preferences. Early methods in this area rely on reward-based RLHF, where human preferences are used to train a reward model that guides reinforcement learning to optimize the LLM’s outputs \cite{christiano2017deep,ziegler2019fine,stiennon2020learning,ouyang2022training}.  
In contrast, reward-free methods such as Direct Preference Optimization (DPO) \cite{rafailov2024direct} and its variants \cite{azar2024general,ethayarajh2024kto,park2024disentangling,meng2024simpo} bypass the explicit learning of a reward model. These approaches directly optimize the model based on pre-collected human preference data through pairwise comparisons, avoiding the construction of an explicit reward function. This often leads to improved data efficiency and scalability for LLM fine-tuning.  
Moreover, recent studies \cite{guo2024direct,pang2024iterative,xiong2024iterative,ye2024online,shani2024multi,cen2025value,zhangself,das2025active} propose online variants of RLHF, where preference data are collected interactively from LLM annotators to iteratively evaluate and update the current policy. These works can be viewed as reward-based preference optimization relying on online collective preference datasets. 

\noindent\textbf{Self-improvement of LLMs.}  
Self-improvement of large language models is an emerging paradigm aiming to enhance model capabilities by leveraging the model’s own outputs as training signals. A prominent line of work is based on RLHF: starting from a supervised fine-tuned (SFT) model, \citet{sun2023salmon} prompt the SFT model to generate preference labels by selecting preferred responses according to certain principles, then train a principle-driven reward model and optimize the policy via PPO. \citet{yuan2024self} build a preference dataset from their own SFT model fine-tuned on instruction-following and evaluation data, followed by DPO training. \citet{chenbootstrapping} focus on further enhancing a DPO-tuned model through bootstrapping with implicit rewards. Another related direction is Best-of-$N$ distillation \cite{sessabond,yang2025faster}, where a model learns from its own Best-of-$N$ sampled outputs to improve consistency and generalization.

\section{Conclusion}
We propose Preference-based Policy Optimization (PbPO), a framework for bootstrapping large language models (LLMs) by iteratively refining both the policy and reward model via a min-max optimization game. Unlike conventional RLHF methods, which can suffer from reward misspecification and premature convergence, PbPO leverages confidence set constraints to ensure robust and reliable policy improvement. Guided exploration further enhances data collection by maintaining uncertainty, enabling continual self-improvement. Our theoretical analysis provides high-probability guarantees, and extensive experiments on multiple benchmarks show that PbPO consistently outperforms state-of-the-art preference optimization methods.

\section{Acknowledgments}
We thank the anonymous reviewers for their helpful comments and suggestions. This work was supported by SI-TECH Information Technology Co., Ltd. 

\bibliography{aaai2026}
\appendix
\onecolumn

\section{Contents of Appendix}
\begin{enumerate}
	\item[] {\bf Proofs for PbPO with Sequence-Level Reward Model} \dotfill 10
	\begin{enumerate}
		\item[] {Proposition 1 (Confidence)} \dotfill 10
		\item[] {Proof of Theorem 1 (Regret bound with sequence-level RM)} \dotfill 10
	    \item[] {Proof of Theorem 2 (Regret lower bound)} \dotfill 14
	\end{enumerate}
	\item[] {\bf Proofs for PbPO with Token-Level Reward Model} \dotfill 15
    \begin{enumerate}
	\item[] {Proposition 2 (Confidence)} \dotfill 15
	\item[] {Proof of Theorem 3 (Regret bound with token-level RM)} \dotfill 16
	\item[] {Proof of Theorem 4 (Regret lower bound)} \dotfill 18
    \end{enumerate}
    \item[] {\bf Supporting Lemmas} \dotfill 20
    \begin{enumerate}
    	\item[] {Lemma 1 (Bounding the expected $L_1$-norm of reward difference)} \dotfill 20
    	\item[] {Definiton 1 ($\epsilon$-bracketing number of sequence-level preference objectives)} \dotfill 21
    	\item[] {Proposition 3 (Bounding bracketing number of sequence-level preference objectives)} \dotfill 21
    	 \item[] {Definition 2 ($\epsilon$-bracketing number for token-level preference objectives)} \dotfill 22
    	 \item[] {Proposition 4 (Bounding bracketing number of token-level preference objectives)} \dotfill 22
    	 \item[] {Lemma 2 (Martingale exponential inequality (Theorem 13.2 of \cite{zhang2023mathematical}))} \dotfill 22
    	 \item[] {Lemma 3 (Uniform convergence for preference with bracketing number))} \dotfill 23
    	 \item[] {Lemma 4 (Concentration of inverse covariances \cite{zanette2021cautiously})} \dotfill 24    	 
    	 \item[] {Lemma 5 (Performance of preference optimization)} \dotfill 24     	 
    \end{enumerate}
    \item[] {\bf Technical Lemmas} \dotfill 24
    \begin{enumerate}
    	\item[] {Proposition 5 (Covering number of the bounded vector space)} \dotfill 24     
    	\item[] {Proposition 6 (Covering number of Cartesian producting vector space)} \dotfill 24   
    	\item[] {Lemma 6 (Elliptical potential lemma \cite{dani2008stochastic,abbasi2011improved})} \dotfill 25
    \end{enumerate}
\end{enumerate}

\section{Proofs for PbPO with Sequence-Level Reward Model} \label{apdx:seq}

\begin{proposition}[Confidence] \label{pro:conf}
	For any $\delta \in (0, 1]$, let $\zeta_k = \mathcal{O}(d\log(Bk/\delta))$, then under Assumptions \ref{asp:linearboundseq} \& \ref{asp:reaseq}, we have
	\begin{align*}
		\mathbb{P}(\forall k \in [K], r^\star \in \mathcal{R}(\mathcal{D}^{\rm pref}_k)) \geq 1 - \delta.
	\end{align*}
\end{proposition}

\begin{proof}
	Using Lemma \ref{lemma:mleper} with Proposition \ref{prop:bracketing} for linear RM, we have for any $k \in [K]$, with probability at least $1-\delta_k$, the event $\mathcal{E}_k: r^\star \in \mathcal{R}(\mathcal{D}^{\rm pref}_k)$ holds because
	\begin{align*}
		\sum_{s=1}^{k} \log \mathbb{P}_{\hat{r}_k}(o_s\mid \tau^0_s, \tau^1_s) \leq& \sum_{s=1}^{k} \log \mathbb{P}_{r^\star}(o_s\mid \tau^0_s, \tau^1_s) + c (d\log(Bk) + \log(k/\delta_k)) \\
		\leq&\sum_{s=1}^{k} \log \mathbb{P}_{r^\star}(o_s\mid \tau^0_s, \tau^1_s) + \mathcal{O}\left( d\log(Bk/\delta_k) \right),
	\end{align*}
	where $\hat{r}_k = \mathop{\arg\max}_{r \in \mathcal{G}_r^{\rm seq}} \sum_{s=1}^{k} \log \mathbb{P}_{r^\star}(o_s\mid \tau^0_s, \tau^1_s)$. We assign the probability $\delta_k = (6/\pi^2)/k^2\delta$ for the $k$-th episode, which, using the above inequalities from $k=1$ to $K$ with an union bound, we obtain the results of Proposition \ref{pro:conf}.
\end{proof}

\begin{proof}[\bf Proof of Theorem \ref{thm:regretseq}]
	We denote $r_{\pi^k} = \mathop{\arg\min}_{r\in\mathcal{R}(\mathcal{D}^{\rm pref}_k)} {\left(J(\pi^k, r) - J(\pi_{\rm ref}^k, r)\right)}$ and $r_{\pi^\star} = \mathop{\arg\min}_{r\in\mathcal{R}(\mathcal{D}^{\rm pref}_k)} {\left( J({\pi^\star}, r) - J(\pi_{\rm ref}^k, r)\right)}$. Then, under Assumption \ref{asp:reaseq}, a regret bound holds with probability at least $1-\delta$,
	\begin{align*}
		{\rm Regret}(K):=&\sum_{k=1}^K \left( J(\pi^\star, r^\star) - J(\pi^k, r^\star) \right) \\
		=& \sum_{k=1}^K \left(J(\pi^\star, r^\star) - J(\pi_{\rm ref}^k, r^\star)  - \left( J(\pi^k, r^\star) - J(\pi_{\rm ref}^k, r^\star) \right) \right) \\
		\leq& \sum_{k=1}^K \Big[ J(\pi^\star, r^\star) - J(\pi_{\rm ref}^k, r^\star) - \left( J(\pi^\star, r_{\pi^\star}) - J(\pi_{\rm ref}^k, r_{\pi^\star})  \right) - \left(  J(\pi^k, r^\star)  -  J(\pi_{\rm ref}^k, r^\star)  - \left(  J({\pi}^k, r_{\pi^k} ) - J(\pi_{\rm ref}^k, r_{\pi^k})  \right)    \right) \Big] \\
		\leq&  \sum_{k=1}^K \Big[  J(\pi^\star, r^\star) - J(\pi_{\rm ref}^k, r^\star) - \left( J(\pi^\star, r_{\pi^\star}) - J(\pi_{\rm ref}^k, r_{\pi^\star})  \right)    \Big] \\
		=& \sum_{k=1}^K \mathbb{E}_{\tau^0 \sim \pi^\star, \tau^1 \sim \pi_{\rm ref}^k} \left[r^\star(x,y^0) - r^\star(x,y^1) - \left( r_{\pi^\star}(x, y^0) - r_{\pi^\star}(x, y^1)  \right) \right],
	\end{align*}
	where the second step follows from the fact that $\pi^k = \mathop{\arg\max}_{\pi \in \Pi} \min_{r \in \mathcal{R}(\mathcal{D}^{\rm pref}_{k})} \left(J(\pi, r) - J(\pi_{\rm ref}^k, r)\right)$, the third step comes from the fact that $r_{\pi^k} = \mathop{\arg\min}_{r\in\mathcal{R}(\mathcal{D}^{\rm pref}_{k})} {\left(J(\pi^k, r) - J(\pi_{\rm ref}^k, r)\right)}$ and $r^\star \in \mathcal{R}(\mathcal{D}^{\rm pref}_{k})$ for all $k \in [K]$ with probability at least $1-\delta$ (by Proposition \ref{pro:conf}). 
	
	Under Assumptions \ref{asp:linearboundseq} \& \ref{asp:reaseq}, we suppose that the reward model $r^\star$ and $r_{\pi^\star}$ can be represented as linear models with parameters $\theta^\star$ and $\theta_{\pi^\star}$, respectively. Then we have
	\begin{align*} 
		{\rm Regret}(K) \leq& \sum_{k=1}^K \mathbb{E}_{\tau^0 \sim \pi^\star, \tau^1 \sim \pi_{\rm ref}^k} \left[r^\star(x,y^0) - r^\star(x,y^1) - \left( r_{\pi^\star}(x, y^0) - r_{\pi^\star}(x, y^1)  \right) \right] \\
		=& \sum_{k=1}^K \mathbb{E}_{\tau^0 \sim \pi^\star, \tau^1 \sim \pi_{\rm ref}^k} \left[r^\star(\tau^0) - r^\star(\tau^1)  - \left( r_{\pi^\star}(\tau^0) - r_{\pi^\star}(\tau^1) \right) \right] \quad {(denote\ r(\tau) := r(x,y)\ for\ short)} \\
		=& \sum_{k=1}^K  \mathbb{E}_{\tau^0 \sim \pi^\star, \tau^1 \sim \pi_{\rm ref}^k}  \left[ (\theta^\star - \theta_{\pi^\star})^\top (\phi(\tau^0) - \phi(\tau^1) \right] \\
		=& \sum_{k=1}^K  \mathbb{E}_{\tau^0 \sim \pi^\star, \tau^1 \sim \pi_{\rm ref}^k} \left[ (\theta^\star - \theta_{\pi^\star})^\top (\phi(\tau^0) - \phi(\tau^1) \right] \mathbf{1} \left\{ \Big\Vert  \mathbb{E}_{\tau^0 \sim \pi^\star, \tau^1 \sim \pi_{\rm ref}^k} \left[  \phi(\tau^0) - \phi(\tau^1)  \right]  \Big\Vert_{\Sigma_k^{-1}} \leq 1 \right\} \\
		& + \sum_{k=1}^K  \mathbb{E}_{\tau^0 \sim \pi^\star, \tau^1 \sim \pi_{\rm ref}^k} \left[ (\theta^\star - \theta_{\pi^\star})^\top (\phi(\tau^0) - \phi(\tau^1) \right] \mathbf{1} \left\{ \Big\Vert  \mathbb{E}_{\tau^0 \sim \pi^\star, \tau^1 \sim \pi_{\rm ref}^k} \left[  \phi(\tau^0) - \phi(\tau^1)  \right] \Big\Vert_{\Sigma_k^{-1}} > 1 \right\} \\
		=:& ({\rm I}) + ({\rm II}),
	\end{align*}
	where $\mathbf{1}\{A\}$ is an indicator function of event $A$, and the PSD covariance matrix $\Sigma_k$ is defined as:
	\begin{align*}
		\Sigma_k := \lambda I + \sum_{s=1}^{k-1} \mathop{\mathbb{E}}\limits_{\tau^0 \sim \hat{\pi}^s, \tau^1 \sim \pi_{\rm ref}^s} \left[(\phi(\tau^0) - \phi(\tau^1)) (\phi(\tau^0) - \phi(\tau^1))^\top \right],
	\end{align*}
	where $\lambda > 0$ is a coefficient to be specified later.
	
	We first bound ${\rm (II)}$ as follows:
	\begin{align*}
		({\rm II}) =&\sum_{k=1}^K  \mathbb{E}_{\tau^0 \sim \pi^\star, \tau^1 \sim \pi_{\rm ref}^k} \left[(\theta^\star - \theta_{\pi^\star})^\top (\phi(\tau^0) - \phi(\tau^1)) \right] \mathbf{1} \left\{ \Big\Vert \mathop{\mathbb{E}}_{\tau^0 \sim \pi^\star, \tau^1 \sim \pi_{\rm ref}^k} \left[  \phi(\tau^0) - \phi(\tau^1)  \right] \Big\Vert_{\Sigma_k^{-1}} > 1 \right\} \\
		\leq& 4B \sum_{k=1}^K \mathbf{1} \left\{ \Big\Vert \mathbb{E}_{\tau^0 \sim \pi^\star, \tau^1 \sim \pi_{\rm ref}^k} \left[  \phi(\tau^0) - \phi(\tau^1)  \right] \Big\Vert_{\Sigma_k^{-1}} > 1 \right\} \\
		\leq& 4B \sqrt{K} \sqrt{\sum_{k=1}^K \min \Big\{1,   \Big\Vert \mathbb{E}_{\tau^0 \sim \pi^\star, \tau^1 \sim \pi_{\rm ref}^k} [ \phi(\tau^0) - \phi(\tau^1)] \Big\Vert_{\Sigma_k^{-1}}^2 \Big\}},
	\end{align*}
	where the first inequality follows from Assumption 1, the second inequality uses the Cauchy–Schwarz inequality.
	
	Next, we bound ${\rm (I)}$ using the Cauchy–Schwarz inequality as follows:
	\begin{align*}
		&\sum_{k=1}^K \mathbb{E}_{\tau^0 \sim \pi^\star, \tau^1 \sim \pi_{\rm ref}^k} \left[ (\theta^\star - \theta_{\pi^\star})^\top (\phi(\tau^0) - \phi(\tau^1)) \right] \mathbf{1} \left\{ \Big\Vert \mathbb{E}_{\tau^0 \sim \pi^\star, \tau^1 \sim \pi_{\rm ref}^k} \left[  \phi(\tau^0) - \phi(\tau^1)  \right] \Big\Vert_{\Sigma_k^{-1}} \leq 1 \right\} \\
		\leq& \sum_{k=1}^K \left\Vert \theta^\star -  \theta_{\pi^\star} \right\Vert_{\Sigma_k} \Big\Vert \mathbb{E}_{\tau^0 \sim \pi^\star, \tau^1 \sim \pi_{\rm ref}^k} [ \phi(\tau^0) - \phi(\tau^1)] \Big\Vert_{\Sigma_k^{-1}} \mathbf{1} \left\{ \Big\Vert \mathbb{E}_{\tau^0 \sim \pi^\star, \tau^1 \sim \pi_{\rm ref}^k}  \left[  \phi(\tau^0) - \phi(\tau^1)  \right] \Big\Vert_{\Sigma_k^{-1}} \leq 1 \right\} \\
		\leq& \sum_{k=1}^K \Vert  \theta^\star -  \theta_{\pi^\star} \Vert_{\Sigma_k}  \min \Big\{1,   \Big\Vert \mathbb{E}_{\tau^0 \sim \pi^\star, \tau^1 \sim \pi_{\rm ref}^k} [ \phi(\tau^0) - \phi(\tau^1) ] \Big\Vert_{\Sigma_k^{-1}} \Big\} \\
		\leq& \sqrt{ \sum_{k=1}^K \Vert  \theta^\star -  \theta_{\pi^\star} \Vert_{\Sigma_k}^2 }  \sqrt{ \sum_{k=1}^K \min \Big\{  1,   \Big\Vert \mathbb{E}_{\tau^0 \sim \pi^\star, \tau^1 \sim \pi_{\rm ref}^k} [ \phi(\tau^0) - \phi(\tau^1) ] \Big\Vert_{\Sigma_k^{-1}}^2 \Big\}}.
	\end{align*}
	
	Then coming back to the regret bound, we have
	\begin{align*}
		{\rm Regret}(K) \leq&\sum_{k=1}^K \mathbb{E}_{\tau^0 \sim \pi^\star, \tau^1 \sim \pi_{\rm ref}^k} \left[r^\star(\tau^0) - r^\star(\tau^1)  - \left( r_{\pi^\star}(\tau^0) - r_{\pi^\star}(\tau^1) \right) \right]  \\
		\leq& \left( \sqrt{ \sum_{k=1}^K \Vert \theta^\star -  \theta_{\pi^\star} \Vert_{\Sigma_k}^2 } + 4B\sqrt{K} \right) \sqrt{ \sum_{k=1}^K \min \Big\{1,   \Big\Vert \mathbb{E}_{\tau^0 \sim \pi^\star, \tau^1 \sim \pi_{\rm ref}^k} [ \phi(\tau^0) - \phi(\tau^1) ] \Big\Vert_{\Sigma_k^{-1}}^2 \Big\}}.
	\end{align*}
	
	In the following, we will upper bound the RHS in two steps.
	
	\noindent\textbf{Step 1.} We bound $\sqrt{ \sum_{k=1}^K \Vert  \theta^\star  - \theta_{\pi^\star} \Vert_{\Sigma_k}^2 }$ as follows:
	\begin{align} 
		\begin{split}
			\sqrt{ \sum_{k=1}^K \Vert  \theta^\star - \theta_{\pi^\star} \Vert_{\Sigma_k}^2} =& \sqrt{\sum_{k=1}^K\left( \lambda \Vert \theta^\star - \theta_{\pi^\star} \Vert_2^2 + \sum_{s=1}^{k-1}\mathop{\mathbb{E}}\limits_{\tau^0 \sim\hat{\pi}^s, \tau^1 \sim \pi_{\rm ref}^s} \left[\vert (\theta^\star - \theta_{\pi^\star})^\top (\phi(\tau^0) - \phi(\tau^1) \vert^2 \right] \right) } \\
			{\leq}& 2B\sqrt{ \lambda K} + \sqrt{\sum_{k=1}^K \sum_{s=1}^{k-1}\mathop{\mathbb{E}}\limits_{\tau^0 \sim \hat{\pi}^s, \tau^1 \sim \pi_{\rm ref}^s} \left[\vert (\theta^\star - \theta_{\pi^\star})^\top  (\phi(\tau^0) - \phi(\tau^1)) \vert^2 \right]} \\
			=& 2B\sqrt{ \lambda K} + \sqrt{\sum_{k=1}^{K-1} \sum_{s=1}^{k}\mathop{\mathbb{E}}\limits_{\tau^0 \sim \hat{\pi}^s, \tau^1 \sim \pi_{\rm ref}^s} \left[\vert (\theta^\star - \theta_{\pi^\star})^\top (\phi(\tau^0) - \phi(\tau^1)) \vert^2 \right]} ,
		\end{split}
	\end{align}
	where the inequality follows from Assumption 1 and the fact that $\sqrt{a+b} \leq \sqrt{a} + \sqrt{b}$ for any $a,b \geq 0$.
	
	Using Lemma \ref{lemma:l1logeexp} on sequence-level RM class $\mathcal{G}_r^{\rm seq}$ by iteratively setting $\pi_0 = \hat{\pi}^s$, $\pi_1 = \pi_{\rm ref}^s$ over $s \in \{1,2,\ldots, k\}$, we have
	\begin{align*} 
		\sum_{s=1}^{k}\mathop{\mathbb{E}}\limits_{\tau^0_{s} \sim \hat{\pi}^s, \tau^1 \sim \pi_{\rm ref}^s} \left[\vert (\theta^\star - \theta_{\pi^\star})^\top (\phi(\tau^0) - \phi(\tau^1)) \vert^2 \right] \leq - \kappa^2 \sum_{s=1}^k \log \mathbb{E}_{\tau^0 \sim \hat{\pi}^s, \tau^1 \sim  \pi_{\rm ref}^s \atop o \sim \mathbb{P}_{r^\star}(\cdot \mid \tau^0, \tau^1)} \exp\left( \ell_{r_{\theta_{\pi^\star}}}(o; \tau^0, \tau^1) \right).
	\end{align*}
	
	Using Lemma \ref{lemma:unformart} on the sequence-level RM setting by bounding the bracketing number with Proposition \ref{prop:bracketing}, we  iteratively set $\pi_0 = \hat{\pi}^s$, $\pi_1 = \pi_{\rm ref}^s$ over $s \in \{1,2,\ldots, k\}$. We have for any $\delta_k \in (0,1]$, with probability at least $1-\sum_{k=1}^K\delta_k$, 
	\begin{align*}
		\forall k \in [K],\ \forall \theta \in \Theta,\ -\sum_{s=1}^{k}\log \mathbb{E}_{\tau^0 \sim \hat{\pi}^s, \tau^1 \sim \pi_{\rm ref}^s \atop o \sim \mathbb{P}_{r^\star}(\cdot \mid \tau^0, \tau^1)} \exp\left( \ell_{r_\theta}(o; \tau^0, \tau^1) \right) \leq c (d\log(Bk) + \log(1/\delta_k)),
	\end{align*}
	with some constant $c>0$ and $\Theta := \{\theta \in \mathbb{R}^d: \Vert \theta \Vert_2 \leq B\}$ denote the RM parameter class in Assumption \ref{asp:linearboundseq}.
	
	Thus we have
	\begin{align*}
		\forall k \in [K],\ \forall \theta \in \Theta,\  \sum_{s=1}^{k}\mathop{\mathbb{E}}\limits_{\tau^0_{s} \sim \hat{\pi}^s, \tau^1 \sim \pi_{\rm ref}^s} \left[\vert (\theta^\star - \theta_{\pi^\star}) (\phi(\tau^0) - \phi(\tau^1))^\top \vert^2 \right] \leq c \kappa^2  (d\log(Bk) + \log(1/\delta_k)).
	\end{align*}
	
	By assigning the probability $\delta_k = (6/\pi^2)/k^2\delta$ for the $k$-th episode, which, using the above inequalities from $k=1$ to $K$ with an union bound, we obtain the result of {\bf step 1} such that with probability at least $1-\delta$,
	\begin{align*}
		\sqrt{ \sum_{k=1}^{K-1} \Vert  \theta_{\pi^\star} - \theta^\star \Vert_{\Sigma_k}^2} \leq  2B\sqrt{ \lambda K} + \sqrt{\sum_{k=1}^{K-1} c(1+e^{2B})^2  (d\log(Bk) + \log(k/\delta)) }.
	\end{align*} 
	
	\noindent\textbf{Step 2.} Bounding $ \sqrt{\sum_{k=1}^K \min \Big\{  1,   \Big\Vert \mathbb{E}_{\tau^0 \sim \pi^\star, \tau^1 \sim \pi_{\rm ref}^k} [ \phi(\tau^0) - \phi(\tau^1) ] \Big\Vert_{\Sigma_k^{-1}}^2 \Big\}}$:
	
	Denote the empirical covariance matrix $\hat{\Sigma}_k$ by sampling $\tau^0_s \sim \hat{\pi}_{s}$, $\tau^1_s \sim \pi_{\rm ref}^s$ for each $s \in \{1,2,\ldots,k\}$ as follows:
	\begin{align*}
		\hat{\Sigma}_k := \lambda I + \sum_{s=1}^{k-1} \left[(\phi(\tau^0_{s}) - \phi(\tau^1_{s})) (\phi(\tau^0_{s}) - \phi(\tau^1_{s}))^\top \right].
	\end{align*}
	
	For each $k \in \{1,2\ldots,K\}$, we use Lemma \ref{lemma:inversecov} by setting $\lambda = c d \log(k/\delta_k)$, then we have for any $\delta \in (0,1]$ and an universal constant $c>0$, with probability at least $1-\delta_k$,
	\begin{align*}
		\Big\Vert \mathop{\mathbb{E}}\limits_{\tau^0 \sim {\pi}^\star, \tau^1 \sim \pi_{\rm ref}^s} [ \phi(\tau^0) - \phi(\tau^1) ]  \Big\Vert_{\Sigma_k^{-1}}^2 \leq \frac{5}{3} \Big\Vert  \mathop{\mathbb{E}}\limits_{\tau^0 \sim {\pi}^\star, \tau^1 \sim \pi_{\rm ref}^s}  [ \phi(\tau^0) - \phi(\tau^1) ] \Big\Vert_{\hat{\Sigma}_k^{-1}}^2.
	\end{align*}
	
	Then following from the definition of the enhancer in Algorithm \ref{alg:onlinekd}: $\hat{\pi}^k = \mathop{\arg\max}_{\pi \in \Pi} \Big\Vert \mathop{\mathbb{E}}_{\tau^0 \sim \pi, \tau^1 \sim \pi_{\rm ref}^k} [ \phi(\tau^0) - \phi(\tau^1) ] \Big\Vert_{\hat{\Sigma}_k^{-1}}^2$ and the confidence set, we have:
	\begin{align*}
		\Big\Vert \mathop{\mathbb{E}}_{\tau^0 \sim \pi^\star, \tau^1 \sim \pi_{\rm ref}^k} [ \phi(\tau^0) - \phi(\tau^1)] \Big\Vert_{\hat{\Sigma}_k^{-1}}^2 \leq \Big\Vert \mathop{\mathbb{E}}_{\tau^0 \sim \hat{\pi}^k, \tau^1 \sim \pi_{\rm ref}^k} [ \phi(\tau^0) - \phi(\tau^1) ] \Big\Vert_{\hat{\Sigma}_k^{-1}}^2.
	\end{align*}
	
	Then we use Lemma \ref{lemma:inversecov} by setting $\lambda = c d\log(k/\delta'_k)$ for any $\delta'_k \in (0,1]$ and an universal constant $c>0$, with probability at least $1-\delta'_k$,
	\begin{align*}
		\Big\Vert \mathop{\mathbb{E}}_{\tau^0 \sim \hat{\pi}^k, \tau^1 \sim \pi_{\rm ref}^k} [ \phi(\tau^0) - \phi(\tau^1) ] \Big\Vert_{\hat{\Sigma}_k^{-1}}^2  \leq 3 \Big\Vert \mathop{\mathbb{E}}_{\tau^0 \sim \hat{\pi}^k, \tau^1 \sim \pi_{\rm ref}^k} [ \phi(\tau^0) - \phi(\tau^1) ] \Big\Vert_{\Sigma_k^{-1}}^2 
	\end{align*}
	
	Define 
	\begin{align*}
		\widetilde{\Sigma}_k = \lambda I + \sum_{s=1}^{k-1} \mathop{\mathbb{E}}_{\tau^0 \sim \hat{\pi}^s, \tau^1 \sim \pi_{\rm ref}^s} \left[\phi(\tau^0) - \phi(\tau^1)\right] \mathop{\mathbb{E}}_{\tau^0 \sim \hat{\pi}^s, \tau^1 \sim \pi_{\rm ref}^s} \left[\phi(\tau^0) - \phi(\tau^1) \right]^\top,
	\end{align*}
	we observe that ${\Sigma}_k = \widetilde{\Sigma}_k + \sum_{s=1}^{k-1} {\rm Var}(\phi(\tau^0) - \phi(\tau^1)) \succeq \widetilde{\Sigma}_k \succ 0$. This imples that ${\Sigma}_k^{-1} \preceq \widetilde{\Sigma}_k^{-1}$. Then, we have
	\begin{align*}
		\Big\Vert \mathop{\mathbb{E}}_{\tau^0 \sim \hat{\pi}^k, \tau^1 \sim \pi_{\rm ref}^k} [ \phi(\tau^0) - \phi(\tau^1) ] \Big\Vert_{\Sigma_k^{-1}}^2  \leq \Big\Vert \mathop{\mathbb{E}}_{\tau^0 \sim \hat{\pi}^k, \tau^1 \sim \pi_{\rm ref}^k} [ \phi(\tau^0) - \phi(\tau^1) ] \Big\Vert_{\widetilde{\Sigma}_k^{-1}}^2 
	\end{align*}
	
	We assign the probability $\delta_k = \delta'_k = (3/\pi^2)/k^2\delta$ for the $k$-th episode, which, using the above inequalities from $k=1$ to $K$ with an union bound, we have with probability at least $1-\delta$:
	\begin{align*}
		\sqrt{\sum_{k=1}^K \min \Big\{  1,   \Big\Vert \mathop{\mathbb{E}}_{\tau^0 \sim \pi^\star, \tau^1 \sim \pi_{\rm ref}^k} [ \phi(\tau^0) - \phi(\tau^1) ] \Big\Vert_{\Sigma_k^{-1}}^2 \Big\}} \leq \sqrt{5\sum_{k=1}^K \min \Big\{1,   \Big\Vert \mathop{\mathbb{E}}_{\tau^0 \sim \hat{\pi}^k, \tau^1 \sim \pi_{\rm ref}^k} [\phi(\tau^0) - \phi(\tau^1)] \Big\Vert_{\widetilde{\Sigma}_k^{-1}}^2 \Big\}}
	\end{align*}
	
	Follows from the elliptical potential lemma (Lemma \ref{lemma:epl}), we have 
	\begin{align*}
		\sum_{k=1}^K \min \Big\{1,   \Big\Vert \mathop{\mathbb{E}}_{\tau^0 \sim \hat{\pi}^k, \tau^1 \sim \pi_{\rm ref}^k} [ \phi(\tau^0) - \phi(\tau^1) ] \Big\Vert_{\widetilde{\Sigma}_k^{-1}}^2 \Big\} \leq 2d\log\left( 1 + \frac{4K}{d\lambda} \right).
	\end{align*}
	
	Set $\lambda = c d \log(K/\delta)$ with some constant $c>0$, we finish the {\bf step 2} with probability at least $1-\delta$:
	\begin{align*} 
		\sqrt{\sum_{k=1}^K \min \Big\{  1,   \Big\Vert \mathop{\mathbb{E}}_{\tau^0 \sim \pi^\star, \tau^1 \sim \pi_{\rm ref}^k} [ \phi(\tau^0) - \phi(\tau^1) ] \Big\Vert_{\Sigma_k^{-1}}^2 \Big\}} \leq \sqrt{2d\log\left( 1 + \frac{4K}{cd^2 \log(K/\delta)} \right)}
	\end{align*}
	
	By the results of {\bf step 1} and {\bf step 2}. We have for any $\delta \in (0,1]$, with probability at least $1-2\delta$:
	\begin{align*}
		&{\rm Regret}(K) \\
		\leq&  \left(\sqrt{\sum_{k=1}^{K-1} c\kappa^2  (d\log(Bk) + \log(k/\delta)) } + (4+2\sqrt{c d \log(K/\delta)}) B \sqrt{K} \right)\sqrt{2d\log\left( 1 + \frac{4K}{cd^2 \log(K/\delta)} \right)} \\
		\leq& c_1B\kappa d \sqrt{K}\log(BK/\delta) \sqrt{\log\left( 1 + \frac{4K}{c_2d^2 \log(K/\delta)} \right)},
	\end{align*}
	
	where $c_1,c_2  > 0$ denote some universal constants. This completes the proof.
\end{proof}

	\begin{proof}[\bf Proof of Theorem \ref{thm:lowerseq}]
		We construct a hard instance on the reward function $r^\star(x,y) = \langle \theta^\star, \phi(x,y)\rangle$. Let the feature space be $\{-\frac{1}{\sqrt{d}},\frac{1}{\sqrt{d}}\}^d$ and the parameters be $\theta^\star \in \{-\Delta, \Delta\}^d$, where $\Delta = \sqrt{d/K}/2$
		It can be verified that
		\begin{align*}
			\sup_{x,y}\Vert \phi(x,y)\Vert_2 \leq \max \left\{  \sup_{\phi \in \{-1/\sqrt{d},1/\sqrt{d}\}^d} \Vert \phi \Vert_2, 0 \right\} = 1,
		\end{align*}
        and the assumption $K \geq d^2/(4 B^2)$ imples that
		\begin{align*}
			\Vert \theta^\star \Vert_2 \leq \sup_{\theta \in \{-\Delta, \Delta\}^d} \Vert \theta \Vert_2 =   \frac{d}{2\sqrt{K}}  \leq B.
		\end{align*}
		
		Then we have
		\begin{align*}
			&\mathbb{E}_{\theta^\star} \left[{\rm Regret}(K)\right] \\
			=& \mathbb{E}_{\theta^\star} \sum_{k=1}^K \mathbb{E}_{\pi^k} \left[ \max_{y \in \mathcal{Y}} \langle  \phi(x,y), \theta^\star \rangle - \langle \phi(x,y),  \theta^\star \rangle \right]  \\
			=& \frac{1}{\sqrt{d}} \cdot \mathbb{E}_{\theta^\star}\left[ \sum_{k=1}^K \sum_{j=1}^d \left(  \operatorname{sign}(\theta^\star[j]) - \phi[k][j]  \right) \theta^\star[j] \right] = \frac{2\Delta}{\sqrt{d}}\cdot \sum_{j=1}^d  \sum_{k=1}^K \mathbb{E}_{\theta^\star} \left[ \mathbf{1}\{ \operatorname{sign}(\phi[k][j]) \neq \operatorname{sign}(\theta^\star[j]) \} \right],
		\end{align*}
		where in the second step, the expectation w.r.t. $(x,y)$ drawn from the policy $\pi^k$ is combined into the expectation $\mathbb{E}_{\theta_\star}[\cdot]$, $\phi[k][j]$ represent the $j$-th coordinate of the feature representation $\phi(x,y)$ at the $k$-th episode.
		
		Let $\theta^\star_{-j}$ denote the vector which differs from $\theta^\star$ at its $j$-th coordinate only, i.e., $\theta^\star_{-j}[j]= - \theta^\star[j]$ and $\theta^\star_{-j}[i] = \theta^\star[i]$ if $i \neq j$.
		Then we have
		\begin{align*}
			\sum_{\theta^\star} \mathbb{E}_{\theta^\star} \left[{\rm Regret}(K)\right]  =& \frac{2 \Delta}{\sqrt{d}} \cdot \sum_{\theta^\star} \sum_{j=1}^d  \sum_{k=1}^K \mathbb{E}_{\theta^\star} \left[ \mathbf{1}\{ \operatorname{sign}(\phi[k][j]) \neq \operatorname{sign}(\theta^\star[j]) \} \right] \\
			=& \frac{2 \Delta}{\sqrt{d}} \cdot \sum_{\theta^\star} \sum_{j=1}^d  \sum_{k=1}^K \mathbb{E}_{\theta^\star_{-j}} \left[ \mathbf{1}\{ \operatorname{sign}(\phi[k][j]) \neq \operatorname{sign}(\theta^\star_{-j}[j]) \} \right] \\
			=& \frac{2 \Delta}{\sqrt{d}} \cdot  \sum_{\theta^\star} \sum_{j=1}^d  \sum_{k=1}^K \left( 1 -  \mathbb{E}_{\theta^\star_{-j}} \left[ \mathbf{1}\{ \operatorname{sign}(\phi[k][j]) \neq \operatorname{sign}(\theta^\star[j]) \} \right] \right)
		\end{align*}
		
		This imples that 
		\begin{align*}
			&2 \sum_{\theta^\star} \mathbb{E}_{\theta^\star} \left[{\rm Regret}(K)\right]  \\
			=& \frac{2 \Delta}{\sqrt{d}} \cdot \sum_{\theta^\star} \sum_{j=1}^d  \sum_{k=1}^K \left( 1 + \mathbb{E}_{\theta^\star} \left[ \mathbf{1}\{ \operatorname{sign}(\phi[k][j]) \neq \operatorname{sign}(\theta^\star[j]) \} \right] -  \mathbb{E}_{\theta^\star_{-j}} \left[ \mathbf{1}\{ \operatorname{sign}(\phi[k][j]) \neq \operatorname{sign}(\theta^\star[j]) \} \right] \right) \\
			=& \frac{2 \Delta}{\sqrt{d}} \cdot \sum_{\theta^\star} \sum_{j=1}^d   \left( K-  \sum_{k=1}^K \left(\mathbb{E}_{\theta^\star_{-j}}\left[ \mathbf{1}\{ \operatorname{sign}(\phi[k][j]) \neq \operatorname{sign}(\theta^\star[j]) \}\right] - \mathbb{E}_{\theta^\star} \left[\mathbf{1}\{ \operatorname{sign}(\phi[k][j]) \neq \operatorname{sign}(\theta^\star[j]) \} \right] \right)\right)
		\end{align*}
		
		We define the event $\mathcal{E}_{k,j}(\theta^\star) = \{\operatorname{sign}(\phi[k][j]) \neq \operatorname{sign}(\theta^\star[j]) \}$ for any $k \in [K]$, $j \in [d]$ and $\theta^\star \in \{-\Delta, \Delta\}^d$,
		\begin{align*}
			2 \sum_{\theta^\star} \mathbb{E}_{\theta^\star} \left[{\rm Regret}(K)\right]  =& \frac{2 \Delta}{\sqrt{d}} \cdot \sum_{\theta^\star} \sum_{j=1}^d   \left( K-  \sum_{k=1}^K \left( \mathbb{P}_{\theta^\star_{-j}}(\mathcal{E}_{k,j}(\theta^\star) )- \mathbb{P}_{\theta^\star} (\mathcal{E}_{k,j}(\theta^\star)) \right)\right) \\
			\geq& \frac{2 \Delta}{\sqrt{d}} \cdot \sum_{\theta^\star} \sum_{j=1}^d   \left( K-  K \sup_{\mathcal{E}\in\{ \mathcal{E}_{k,j}(\theta^\star), \overline{\mathcal{E}}_{k,j}(\theta^\star) \}} \left( \mathbb{P}_{\theta^\star_{-j}}(\mathcal{E}) - \mathbb{P}_{\theta^\star} (\mathcal{E}) \right)\right) \\
			=& \frac{2 \Delta}{\sqrt{d}} \cdot \sum_{\theta^\star} \sum_{j=1}^d   \left( K-  K\cdot {\rm TV} \left( \mathbb{P}_{\theta^\star_{-j}}, \mathbb{P}_{\theta^\star} \right)\right) \\
			\geq& \frac{2 \Delta}{\sqrt{d}} \cdot \sum_{\theta^\star} \sum_{j=1}^d   \left( K-  K \sqrt{ \frac{1}{2}{\rm KL} \left(  \mathbb{P}_{\theta^\star}, \mathbb{P}_{\theta^\star_{-j}} \right) } \right),
		\end{align*}
		where ${\rm TV}(\cdot, \cdot)$ and ${\rm KL}(\cdot, \cdot)$ denote the total variation distance and Kullback–Leibler divergence, respectively, the last step follows from Pinsker’s inequality.
		
		Let $\mathcal{H}_{k-1}$ denote a sigma algebra w.r.t. the history before the $k$-th episode, $\mathcal{H}_{k-1} = \sigma(\{x_1, y^0_1, y^1_1, o_1, \ldots, x_{k-1}, y^0_{k-1}, y^1_{k-1}, o_{k-1}\})$. By the chain rule of relative entropy and the Bernoulli distribution of reward, we can further decompose ${\rm KL} \left(  \mathbb{P}_{\theta^\star}, \mathbb{P}_{\theta^\star_{-j}} \right) $ as 
		\begin{align*}
			{\rm KL} \left(  \mathbb{P}_{\theta^\star}, \mathbb{P}_{\theta^\star_{-j}} \right)  =& \sum_{k=1}^K \mathbb{E}_{\theta^\star} \left[ {\rm KL}(\mathbb{P}_{\theta^\star}(o_k \mid x_k, y^0_k, y^1_k; \mathcal{H}_{k-1}), \mathbb{P}_{\theta^\star_{-j}}(o_k \mid x_k, y^0_k, y^1_k; \mathcal{H}_{k-1})) \right] \\
			=& \sum_{k=1}^K \mathbb{E}_{\theta^\star} \left[ {\rm KL}({\rm Ber}(\sigma(\langle \phi(x_k,y^0_k) - \phi(x_k,y^1_k), \theta^\star\rangle)), {\rm Ber}(\sigma(\langle  \phi(x_k,y^0_k) - \phi(x_k,y^1_k), \theta^\star_{-j} \rangle))) \right] 
		\end{align*}
		
		It then follows from a Taylor expansion and since $\sigma'(x) \leq 1/4$,
		\begin{align*}
			  &{\rm KL}\left({\rm Ber}(\sigma(\langle \phi(x_k,y^0_k) - \phi(x_k,y^1_k), \theta^\star\rangle)), {\rm Ber}(\sigma(\langle  \phi(x_k,y^0_k) - \phi(x_k,y^1_k), \theta^\star_{-j} \rangle)) \right) \\
			  \leq& \frac{1}{8} \langle  \phi(x_k,y^0_k) - \phi(x_k,y^1_k) , \theta^\star - \theta^\star_{-j}  \rangle^2 \leq 2 \cdot \frac{ \Delta^2 } {d}.
		\end{align*}
		
		Then we have
		\begin{align*}
			{\rm KL} \left(  \mathbb{P}_{\theta^\star}, \mathbb{P}_{\theta^\star_{-j}} \right) \leq 2 K \cdot \frac{ \Delta^2 } {d},
		\end{align*}
		and thus
		\begin{align*}
			\sum_{\theta^\star} \mathbb{E}_{\theta^\star} \left[{\rm Regret}(K)\right]  \geq \Delta \sqrt{d} K \sum_{\theta^\star} \left( 1 - \Delta \sqrt{K/d}  \right) 
		\end{align*}
		Further let $\Delta = \sqrt{d/K}/2$, which imples that $ 1 - \Delta \sqrt{K/d}  = 1/2$. We have there exist $\theta^\star \in \{-\Delta,\Delta\}^d$, such that
	
		
		\begin{align*}
			\mathbb{E}_{\theta^\star} \left[{\rm Regret}(K)\right]   \geq \Omega\left(d\sqrt{K}\right),
		\end{align*}
		which completes the proof.
		
	\end{proof}

\section{Proofs for PbPO with Token-Level Reward Model} \label{apdx:token}

\begin{proposition}[Confidence] \label{pro:confaction}
	For any $\delta \in (0, 1]$, let $\zeta_k = \mathcal{O}(dH\log(Bk\sqrt{H}/\delta))$, then under Assumptions \ref{asp:linearboundtoken} \& \ref{asp:reaonlinetoken}, we have
	\begin{align*}
		\mathbb{P}(\forall k \in [K], r^\star \in \mathcal{R}(\mathcal{D}^{\rm pref}_k)) \geq 1 - \delta.
	\end{align*}
\end{proposition}
\begin{proof}
	Using Lemma \ref{lemma:mleper} with Proposition \ref{prop:bracketingaction} for token-level RM, we have for any $k \in [K]$, with probability at least $1-\delta_k$, the event $\mathcal{E}_k:  {r}^\star \in \mathcal{R}(\mathcal{D}^{\rm pref}_k)$ holds because
	\begin{align*}
		\sum_{s=1}^{k} \log \mathbb{P}_{\hat{r}_k}(o_s\mid \tau^0_s, \tau^1_s) \leq& \sum_{s=1}^{k} \log \mathbb{P}_{r^\star}(o_s\mid \tau^0_s, \tau^1_s) + c (dH\log(Bk\sqrt{H}) + \log(k/\delta_k)) \\
		\leq&\sum_{s=1}^{k} \log \mathbb{P}_{r^\star}(o_s\mid \tau^0_s, \tau^1_s) + \mathcal{O}\left( dH\log(Bk\sqrt{H}/\delta_k) \right),
	\end{align*}
	where $\hat{r}_k = \mathop{\arg\max}_{r \in \mathcal{G}_{r}^{\rm tok}} \sum_{s=1}^{k} \log \mathbb{P}_{r^\star}(o_s\mid \tau^0_s, \tau^1_s)$. We assign the probability $\delta_k = (6/\pi^2)/k^2\delta$ for the $k$-th episode, which, using the above inequalities from $k=1$ to $K$ with an union bound, we obtain the results of Proposition \ref{pro:confaction}.
\end{proof}
\begin{proof}[\bf Proof of Theorem \ref{thm:regretupperaction}]
	This proof is similar as the proof of sequence-level RM setting in Theorem \ref{thm:regretseq}, except for the definiton of token-level RM $r \in \mathcal{G}_r^{\rm tok}$, defined as $r_h(s_h, a_h)  = \langle\theta_h, \phi(s_h, a_h) \rangle$. For simplicity in the following derivations, we denote $\theta = (\theta_1, \theta_2, \ldots, \theta_H) \in \mathbb{R}^{dH}$ for any $\{\theta_h\}_{h=1}^H$ and $\phi(\tau) = (\phi(s_1,a_1), \phi(s_2,a_2), \ldots, \phi(s_H, a_H)) \in \mathbb{R}^{dH}$ for any $\tau=(s_1, a_1, \ldots, s_H, a_H)$. By Assumption \ref{asp:linearboundtoken}, we have $\Vert \theta \Vert_2 = \sqrt{\Vert \theta_1 \Vert_2^2 + \ldots + \Vert \theta_H \Vert_2^2} \leq \sqrt{H}B$ and similarly $\Vert \phi(\tau) \Vert_2 \leq \sqrt{H}$. 
	
	We denote $r^{\pi^\star} = \mathop{\arg\min}_{r\in\mathcal{R}(\mathcal{D}^{\rm pref}_k)} {\left(J({\pi^\star}, r) - J(\pi_{\rm ref}^k, r)\right)}$. Then under Assumptions \ref{asp:linearboundtoken} \& \ref{asp:reaonlinetoken}, a regret bound holds with probability at least $1-\delta$:
	\begin{align}\label{eq:onlineactionproof} 
		\begin{split}
		    {\rm Regret}(K) \leq& \sum_{k=1}^K \mathbb{E}_{\tau^0 \sim \pi^\star, \tau^1 \sim \pi_{\rm ref}^k} \left[\sum_{h=1}^H \left( r^\star_h(s^0_h, a^0_h) - r^\star(s^1_h, a^1_h) - \left( r^{\pi^\star}_h(s^0_h, a^0_h) - r^{\pi^\star}_h(s^1_h, a^1_h)  \right) \right) \right] \\
			=& \sum_{k=1}^K \mathbb{E}_{\tau^0 \sim \pi^\star, \tau^1 \sim \pi_{\rm ref}^k} \left[ (\theta^\star  -  \theta^{\pi^\star})^\top (\phi(\tau^0) - \phi(\tau^1)) \right] \\
			=& \sum_{k=1}^K \mathbb{E}_{\tau^0 \sim \pi^\star, \tau^1 \sim \pi_{\rm ref}^k} \left[ (\theta^\star  -  \theta^{\pi^\star})^\top (\phi(\tau^0) - \phi(\tau^1)) \right] \mathbf{1} \left\{ \left\Vert \mathbb{E}_{\tau^0 \sim \pi^\star, \tau^1 \sim \pi_{\rm ref}^k}[\phi(\tau^0) - \phi(\tau^1)] \right\Vert_{\Sigma^{-1}_k} > 1  \right\} \\
			&+ \sum_{k=1}^K \mathbb{E}_{\tau^0 \sim \pi^\star, \tau^1 \sim \pi_{\rm ref}^k} \left[ (\theta^\star  -  \theta^{\pi^\star})^\top (\phi(\tau^0) - \phi(\tau^1)) \right] \mathbf{1} \left\{ \left\Vert \mathbb{E}_{\tau^0 \sim \pi^\star, \tau^1 \sim \pi_{\rm ref}^k}[\phi(\tau^0) - \phi(\tau^1)]  \right\Vert_{\Sigma^{-1}_k} \leq 1  \right\} \\
			=& ({\rm I}) + ({\rm II}),
		\end{split}
	\end{align}
	where $\mathbf{1}\{A\}$ is an indicator function of event $A$, and the PSD covariance matrix $\boldsymbol{\Sigma}_k$ is defined as:
	\begin{align*}
		\boldsymbol{\Sigma}_k := \lambda I + \sum_{s=1}^{k-1} \mathop{\mathbb{E}}\limits_{\tau^0_{s} \sim \hat{\pi}^s, \tau^1_{s} \sim \pi_{\rm ref}^s} \left[(\phi(\tau^0_{s}) - \phi(\tau^1_{s})) (\phi(\tau^0_{s}) - \phi(\tau^1_{s}))^\top \right],
	\end{align*}
	where $\lambda > 0$ is a coefficient to be specified later.
	
	We first bound ${\rm (I)}$ of Eq. (\ref{eq:onlineactionproof}) as follows:
	
	\begin{align*}
		({\rm I}) =& \sum_{k=1}^K \mathbb{E}_{\tau^0 \sim \pi^\star, \tau^1 \sim \pi_{\rm ref}^k} \left[ (\theta^\star  -  \theta^{\pi^\star})^\top (\phi(\tau^0) - \phi(\tau^1)) \right] \mathbf{1} \left\{ \left\Vert \mathbb{E}_{\tau^0 \sim \pi^\star, \tau^1 \sim \pi_{\rm ref}^k}[\phi(\tau^0) - \phi(\tau^1)] \right\Vert_{\boldsymbol{\Sigma}^{-1}_k} > 1  \right\} \\
		\leq& 4BH \sum_{k=1}^K \mathbf{1} \left\{ \left\Vert \mathbb{E}_{\tau^0 \sim \pi^\star, \tau^1 \sim \pi_{\rm ref}^k}[\phi(\tau^0) - \phi(\tau^1)] \right\Vert_{\boldsymbol{\Sigma}^{-1}_k} > 1  \right\} \\
		\leq& 4BH\sqrt{K} \sqrt{\sum_{k=1}^K \min\left\{ 1, \left\Vert \mathbb{E}_{\tau^0 \sim \pi^\star, \tau^1 \sim \pi_{\rm ref}^k}[\phi(\tau^0) - \phi(\tau^1)] \right\Vert_{\boldsymbol{\Sigma}^{-1}_k} \right\}}
	\end{align*}
	
	Next, we bound ${\rm (II)}$ of Eq. (\ref{eq:onlineactionproof}) using Cauchy-Schwarz inequality as follows:
	\begin{align*}
		({\rm II}) =& \sum_{k=1}^K \mathbb{E}_{\tau^0 \sim \pi^\star, \tau^1 \sim \pi_{\rm ref}^k} \left[ (\theta^\star  -  \theta^{\pi^\star})^\top (\phi(\tau^0) - \phi(\tau^1)) \right] \mathbf{1} \left\{ \left\Vert \mathbb{E}_{\tau^0 \sim \pi^\star, \tau^1 \sim \pi_{\rm ref}^k}[\phi(\tau^0) - \phi(\tau^1)]  \right\Vert_{\boldsymbol{\Sigma}^{-1}_k} \leq 1  \right\} \\
		=&  \sum_{k=1}^K \left\Vert \theta^\star  -  \theta^{\pi^\star}  \right\Vert_{\Sigma_k} \left\Vert  \mathbb{E}_{\tau^0 \sim \pi^\star, \tau^1 \sim \pi_{\rm ref}^k} \left[ \phi(\tau^0) - \phi(\tau^1) \right]  \right\Vert_{\boldsymbol{\Sigma}^{-1}_k} \mathbf{1} \left\{ \left\Vert \mathbb{E}_{\tau^0 \sim \pi^\star, \tau^1 \sim \pi_{\rm ref}^k}[\phi(\tau^0) - \phi(\tau^1)]  \right\Vert_{\boldsymbol{\Sigma}^{-1}_k} \leq 1  \right\} \\
		\leq&  \sum_{k=1}^K \left\Vert \theta^\star  -  \theta^{\pi^\star}  \right\Vert_{\Sigma_k} \min \left\{ 1, \left\Vert \mathbb{E}_{\tau^0 \sim \pi^\star, \tau^1 \sim \pi_{\rm ref}^k} \left[ \phi(\tau^0) - \phi(\tau^1) \right]  \right\Vert_{\boldsymbol{\Sigma}^{-1}_k} \right\} \\
		\leq& \sqrt{\sum_{k=1}^K \left\Vert \theta^\star  -  \theta^{\pi^\star}  \right\Vert_{\Sigma_k} } \sqrt{\sum_{k=1}^K \min \left\{ 1, \left\Vert \mathbb{E}_{\tau^0 \sim \pi^\star, \tau^1 \sim \pi_{\rm ref}^k} \left[ \phi(\tau^0) - \phi(\tau^1) \right]  \right\Vert_{\boldsymbol{\Sigma}^{-1}_k} \right\}}
	\end{align*}
	
	Then, using the results of $({\rm I})$ and $({\rm II})$ with Eq. (\ref{eq:onlineactionproof}), we bound the regret as follows
	\begin{align} \label{eq:actionregret}
		{\rm Regret}(K) \leq \left( 4BH\sqrt{K} +  \sqrt{\sum_{k=1}^K \left\Vert \theta^\star  -  \theta^{\pi^\star}  \right\Vert_{\boldsymbol{\Sigma}_k} } \right) \sqrt{\sum_{k=1}^K \min \left\{ 1, \left\Vert \mathbb{E}_{\tau^0 \sim \pi^\star, \tau^1 \sim \pi_{\rm ref}^k} \left[ \phi(\tau^0) - \phi(\tau^1) \right]  \right\Vert_{\boldsymbol{\Sigma}^{-1}_k} \right\}}
	\end{align}
	
	\noindent\textbf{Step 1.} We bound $\sqrt{ \sum_{k=1}^K \Vert  \theta^\star  - \theta_{\pi^\star} \Vert_{\boldsymbol{\Sigma}_k}^2 }$ as follows:
	
	\begin{align} \label{eq:onlinestep1firstaction}
		\begin{split}
			&\sqrt{ \sum_{k=1}^K \Vert  \theta^\star - \theta^{\pi^\star} \Vert_{\boldsymbol{\Sigma}_k}^2} \\
			=& \sqrt{\sum_{k=1}^K \left( \lambda \Vert \theta^\star - \theta_{\pi^\star} \Vert_2^2 + \sum_{s=1}^{k-1}\mathop{\mathbb{E}}\limits_{\tau^0 \sim \hat{\pi}^s, \tau^1 \sim \pi_{\rm ref}^s} \left[\vert (\theta^\star - \theta^{\pi^\star}) (\phi(\tau^0) - \phi(\tau^1)^\top \vert^2 \right] \right) } \\
			{\leq}& 2B\sqrt{ \lambda K H} + \sqrt{\sum_{k=1}^K \sum_{s=1}^{k-1}\mathop{\mathbb{E}}\limits_{\tau^0 \sim \hat{\pi}^s, \tau^1 \sim \pi_{\rm ref}^s} \left[\vert (\theta^\star - \theta^{\pi^\star}) (\phi(\tau^0) - \phi(\tau^1))^\top \vert^2 \right]} \\
			=& 2B\sqrt{ \lambda KH} + \sqrt{\sum_{k=1}^{K-1} \sum_{s=1}^{k}\mathop{\mathbb{E}}\limits_{\tau^0 \sim \hat{\pi}^s, \tau^1 \sim \pi_{\rm ref}^s} \left[\vert (\theta^\star - \theta^{\pi^\star}) (\phi(\tau^0) - \phi(\tau^1))^\top \vert^2 \right]},
		\end{split}
	\end{align}
	where the inequality follows from Assumption \ref{asp:linearboundtoken} and the fact that $\sqrt{a+b} \leq \sqrt{a} + \sqrt{b}$ for any $a,b \geq 0$.
	
	Using Lemma \ref{lemma:l1logeexp} on token-level RM class $\mathcal{G}_r^{\rm tok}$, we have
	\begin{align*}
		\sum_{s=1}^{k}\mathop{\mathbb{E}}\limits_{\tau^0 \sim \hat{\pi}^s, \tau^1 \sim \pi_{\rm ref}^s} \left[\vert (\theta^\star - \theta^{\pi^\star}) (\phi(\tau^0) - \phi(\tau^1))^\top \vert^2 \right] \leq -\kappa^2 \sum_{s=1}^{k} \log \mathbb{E}_{\tau^0 \sim \hat{\pi}^s, \tau^1 \sim \pi_{\rm ref}^s \atop o \sim \mathbb{P}_{r^\star}(\cdot \mid \tau^0, \tau^1)} \exp\left( \ell_{r_{\theta^{\pi^*}}}(o; \tau^0, \tau^1) \right).
	\end{align*}
	
	Using Lemma \ref{lemma:unformart} with Proposition \ref{prop:bracketingaction} on the token-level RM setting. We have for any $\delta_k \in (0,1]$, with probability at least $1-\sum_{k=1}^K \delta_k$,
	\begin{align*}
		\forall k \in [K],\ \forall  \theta \in \Theta^H,\ -\sum_{s=1}^{k} \log \mathbb{E}_{\tau^0 \sim  \hat{\pi}^s, \tau^1 \sim \pi_{\rm ref}^s \atop o \sim \mathbb{P}_{r^\star}(\cdot \mid \tau^0, \tau^1)} \exp\left( \ell_{r_\theta}(o; \tau^0, \tau^1) \right) \leq c\left( dH \log(B\sqrt{H}k) + \log(1/\delta_k) \right),
	\end{align*}
	where $\theta_h \in \Theta = \{\theta_h \in \mathbb{R}^d: \Vert \theta_h \Vert_2 \leq B\}$ denote the RM parameter in Assumption \ref{asp:linearboundtoken}.
	
	Thus we have
	\begin{align*}
		\forall k \in [K],\ \forall  \theta \in \Theta^H, \sum_{s=1}^{k}\mathop{\mathbb{E}}\limits_{\tau^0 \sim \hat{\pi}^s, \tau^1 \sim \pi_{\rm ref}^s} \left[\vert (\theta^\star - \theta^{\pi^\star}) (\phi(\tau^0) - \phi(\tau^1))^\top \vert^2 \right] \leq c \kappa^2 \left( dH \log(B\sqrt{H}k) + \log(1/\delta_k) \right).
	\end{align*}
	
	We come back to Eq. (\ref{eq:onlinestep1firstaction}) by assigning the probability $\delta_k = (6/\pi^2)/k^2\delta$ for the $k$-th episode, which, using the above inequalities from $k=1$ to $K$ with an union bound, we obtain the result of {\bf step 1} such that with probability at least $1-\delta$,
	\begin{align*}
		\sqrt{ \sum_{k=1}^K \Vert  \theta^\star - \theta^{\pi^\star} \Vert_{\boldsymbol{\Sigma}_k}^2} \leq 2B\sqrt{\lambda KH} + \sqrt{\sum_{k=1}^{K-1} c \kappa^2 \left( dH \log(B\sqrt{H}k) + \log(k/\delta) \right)}.
	\end{align*}
	
	\noindent\textbf{Step 2.} We bound $\sqrt{\sum_{k=1}^K \min \left\{ 1, \left\Vert \mathbb{E}_{\tau^0 \sim \pi^\star, \tau^1 \sim \pi_{\rm ref}^k} \left[ \phi(\tau^0) - \phi(\tau^1) \right]  \right\Vert_{\boldsymbol{\Sigma}^{-1}_k} \right\}}$ as follows.
	
	Denote the empirical covariance matrix $\hat{\boldsymbol{\Sigma}}_k$ by sampling $\tau^0_s \sim \hat{\pi}_{s}$, $\tau^1_s \sim \pi_{\rm ref}^s$ for each $s \in \{1,2,\ldots,k\}$ as follows:
	\begin{align*}
		\hat{\boldsymbol{\Sigma}}_k :=& \lambda I + \sum_{s=1}^{k-1} \left[(\phi(\tau^0_{s}) - \phi(\tau^1_{s})) (\phi(\tau^0_{s}) - \phi(\tau^1_{s}))^\top \right]; \\
		\widetilde{\boldsymbol{\Sigma}}_k :=& \lambda I + \sum_{s=1}^{k-1} \mathop{\mathbb{E}}_{\tau^0 \sim \hat{\pi}^s, \tau^1 \sim \pi_{\rm ref}^s} \left[\phi(\tau^0) - \phi(\tau^1)\right] \mathop{\mathbb{E}}_{\tau^0 \sim \hat{\pi}^s, \tau^1 \sim \pi_{\rm ref}^s} \left[\phi(\tau^0) - \phi(\tau^1) \right]^\top.
	\end{align*}
	
	We define the enhancer in Algorithm \ref{alg:onlinekd} as:
	\begin{align*}
		\hat{\pi}^k = \mathop{\arg\max}_{\pi \in \Pi} \Big\Vert \mathop{\mathbb{E}}_{\tau^0 \sim \pi, \tau^1 \sim \pi_{\rm ref}^k} [ \phi(\tau^0) - \phi(\tau^1) ] \Big\Vert_{\hat{\boldsymbol{\Sigma}}_k^{-1}}^2
	\end{align*}
	
	Then using a similar derivation as in the proof of Theorem \ref{thm:regretseq}, we have
	\begin{align*}
		\Big\Vert \mathop{\mathbb{E}}\limits_{\tau^0 \sim {\pi}^\star, \tau^1 \sim \pi_{\rm ref}^s} [ \phi(\tau^0) - \phi(\tau^1) ]  \Big\Vert_{\boldsymbol{\Sigma}_k^{-1}}^2 \leq& \frac{5}{3} \Big\Vert  \mathop{\mathbb{E}}\limits_{\tau^0 \sim {\pi}^\star, \tau^1 \sim \pi_{\rm ref}^s}  [ \phi(\tau^0) - \phi(\tau^1) ] \Big\Vert_{\hat{\boldsymbol{\Sigma}}_k^{-1}}^2 \\
		\leq& \frac{5}{3} \Big\Vert \mathop{\mathbb{E}}_{\tau^0 \sim \hat{\pi}^k, \tau^1 \sim \pi_{\rm ref}^k} [ \phi(\tau^0) - \phi(\tau^1) ] \Big\Vert_{\hat{\boldsymbol{\Sigma}}_k^{-1}}^2 \\
		\leq& 5 \Big\Vert \mathop{\mathbb{E}}_{\tau^0 \sim \hat{\pi}^k, \tau^1 \sim \pi_{\rm ref}^k} [ \phi(\tau^0) - \phi(\tau^1) ] \Big\Vert_{\boldsymbol{\Sigma}_k^{-1}}^2 \\
		\leq&  5 \Big\Vert \mathop{\mathbb{E}}_{\tau^0 \sim \hat{\pi}^k, \tau^1 \sim \pi_{\rm ref}^k} [ \phi(\tau^0) - \phi(\tau^1) ] \Big\Vert_{\widetilde{\boldsymbol{\Sigma}}_k^{-1}}^2,
	\end{align*}
	where the first inequality follows from Lemma \ref{lemma:inversecov} by setting $\lambda = c H d \log(k/\delta_k)$, such that for any $\delta \in (0,1]$ and an universal constant $c>0$, it holds with probability at least $1-\delta_k$. The second inequality follow form the definition of $ \hat{\pi}^k $. The third inequality follow from Lemma \ref{lemma:inversecov} by setting $\lambda = c H d\log(k/\delta'_k)$ for any $\delta'_k \in (0,1]$ and an universal constant $c>0$, such that it holds with probability at least $1-\delta'_k$. The forth inequality follows from the observation that $\boldsymbol{\Sigma}_k = \widetilde{\boldsymbol{\Sigma}}_k + \sum_{s=1}^{k-1} {\rm Var}(\phi(\tau^0) - \phi(\tau^1)) \succeq \widetilde{\boldsymbol{\Sigma}}_k \succ 0$, which imples that $\boldsymbol{\Sigma}_k^{-1} \preceq \widetilde{\boldsymbol{\Sigma}}_k^{-1}$. 
	
	We assign the probability $\delta_k = \delta'_k = (3/\pi^2)/k^2\delta$ for the $t$-th, which, using the above inequalities from $k=1$ to $K$ with an union bound, we have with probability at least $1-\delta$:
	\begin{align*}
		\sqrt{\sum_{k=1}^K \min \Big\{  1,   \Big\Vert \mathop{\mathbb{E}}_{\tau^0 \sim \pi^\star, \tau^1 \sim \pi_{\rm ref}^k} [ \phi(\tau^0) - \phi(\tau^1) ] \Big\Vert_{\boldsymbol{\Sigma}_k^{-1}}^2 \Big\}} \leq \sqrt{5\sum_{k=1}^K \min \Big\{1,   \Big\Vert \mathop{\mathbb{E}}_{\tau^0 \sim \hat{\pi}^k, \tau^1 \sim \pi_{\rm ref}^k} [\phi(\tau^0) - \phi(\tau^1)] \Big\Vert_{\widetilde{\boldsymbol{\Sigma}}_k^{-1}}^2 \Big\}}
	\end{align*}
	
	Follows from the Elliptical Potential Lemma (refer to Lemma \ref{lemma:epl}), we have 
	\begin{align*}
		\sum_{k=1}^K \min \Big\{1,  \Big\Vert \mathop{\mathbb{E}}_{\tau^0 \sim \hat{\pi}^k, \tau^1 \sim \pi_{\rm ref}^k} [ \phi(\tau^0) - \phi(\tau^1) ] \Big\Vert_{\widetilde{\boldsymbol{\Sigma}}_k^{-1}}^2 \Big\} \leq 2Hd\log\left( 1 + \frac{4K}{d\lambda} \right).
	\end{align*}
	
	Set $\lambda = c H d \log(K/\delta)$ with some constant $c>0$, we finish the {\bf step 2} with probability at least $1-\delta$:
	\begin{align*} 
		\sqrt{\sum_{k=1}^K \min \Big\{  1,   \Big\Vert \mathop{\mathbb{E}}_{\tau^0 \sim \pi^\star, \tau^1 \sim \pi_{\rm ref}^k} [ \phi(\tau^0) - \phi(\tau^1) ] \Big\Vert_{\boldsymbol{\Sigma}_k^{-1}}^2 \Big\}} \leq \sqrt{2Hd\log\left( 1 + \frac{4K}{cHd^2 \log(K/\delta)} \right)}
	\end{align*}
	
	By the results of {\bf step 1} and {\bf step 2}. We have for any $\delta \in (0,1]$, with probability at least $1-2\delta$:
	\begin{align*}
		&{\rm Regret}(K) \\
		\leq&  \left(\sqrt{\sum_{k=1}^{K-1} c\kappa^2  (dH\log(B\sqrt{H}k) + \log(k/\delta)) } + (4+2\sqrt{c d \log(K/\delta)}) BH \sqrt{K} \right)\sqrt{2Hd\log\left( 1 + \frac{4K}{cHd^2 \log(K/\delta)} \right)} \\
		\leq& c_1 B \kappa d  H^{3/2} \sqrt{K} \log(B\sqrt{H}K/\delta)\sqrt{\log\left( 1 + \frac{4K}{c_2Hd^2 \log(K/\delta)} \right)},
	\end{align*}
	where $c_1, c_2 > 0$ denote some universal constants.
	This completes the proof.
\end{proof}

\begin{proof}[\bf Proof of Theorem \ref{thm:lowertok}]
	We construct a hard instance on the reward function $r^\star_h(s_h,a_h) = \langle \theta_h^\star, \phi(s_h,a_h)\rangle$. Let the feature space of $\phi_h(s_h,a_h)$ be $\{-\frac{1}{\sqrt{d}},\frac{1}{\sqrt{d}}\}^d$ and the parameters be $\theta^\star_h \in \{-\Delta, \Delta\}^d$ for any $h \in [H]$, where $\Delta = \sqrt{d/K}/2$.
	It can be verified that
	\begin{align*}
		\sup_{s,a}\Vert \phi(s,a)\Vert_2 \leq \max \left\{  \sup_{\phi \in \{-1/\sqrt{d},1/\sqrt{d}\}^d} \Vert \phi \Vert_2, 0 \right\} = 1,
	\end{align*}
	and the assumption $K \geq d^2/(4 B^2)$ imples that
	\begin{align*}
		\Vert \theta^\star_h \Vert_2 \leq \sup_{\theta_h \in \{-\Delta, \Delta\}^d} \Vert \theta_h \Vert_2 =   \frac{d}{2\sqrt{K}}  \leq B.
	\end{align*}
	
	Then, we have
	\begin{align*}
		&\mathbb{E}_{\theta^\star} \left[{\rm Regret}(K)\right] \\
		=& \mathbb{E}_{\theta^\star} \sum_{k=1}^K \mathbb{E}_{\tau^* \sim \pi^\star,\tau \sim \pi^k} \left[ \sum_{h=1}^H \langle  \phi(s^*_h,a^*_h), \theta^\star_h \rangle - \langle \phi(s_h,a_h),  \theta^\star_h \rangle \right]  \\
		=& \frac{1}{\sqrt{d}} \cdot \mathbb{E}_{\theta^\star}\left[ \sum_{k=1}^K \sum_{h=1}^H \sum_{j=1}^d \left(  \operatorname{sign}(\theta^\star_h[j]) - \phi[k][h][j]  \right) \theta^\star_h[j] \right] = \frac{2\Delta}{\sqrt{d}}\cdot \sum_{h=1}^H \sum_{j=1}^d  \sum_{k=1}^K \mathbb{E}_{\theta^\star} \left[ \mathbf{1}\{ \operatorname{sign}(\phi[k][h][j]) \neq \operatorname{sign}(\theta^\star_h[j]) \} \right],
	\end{align*}
	where in the second equality, the expectations w.r.t. $(s_h,a_h)$ drawn from the policy $\pi^k$ and $(s^*_h, a^*_h)$ drawn from the optimal policy $\pi^\star$ are combined into the expectation $\mathbb{E}_{\theta_\star}[\cdot]$, $\phi[k][h][j]$ represent the $j$-th coordinate of feature representation $\phi(s_h,a_h)$ in the $h$-th step decision process at the $k$-th episode.
	
	Let  $\theta^\star_{-hj}$ denote the vector which differs from $\theta^\star$ at its $h$-th step and $j$-th coordinate only, i.e., $\theta^\star_{-hj}[h][j]= - \theta^\star_{h}[j]$ and $\theta^\star_{-j}[h'][i] = \theta^\star_{h'}[i]$ if $i \neq j$ or $h' \neq h$.
	Then using a similar derivation as in the proof of Theorem \ref{thm:lowerseq}, we have  
	\begin{align*}
		&2 \sum_{\theta^\star} \mathbb{E}_{\theta^\star} \left[{\rm Regret}(K)\right]  \\
		=& \frac{2 \Delta}{\sqrt{d}} \cdot \sum_{\theta^\star} \sum_{h=1}^H \sum_{j=1}^d   \left( K-  \sum_{k=1}^K \left(\mathbb{E}_{\theta^\star_{-hj}}\left[ \mathbf{1}\{ \operatorname{sign}(\phi[k][h][j]) \neq \operatorname{sign}({\theta}_h^\star[j]) \}\right] - \mathbb{E}_{\theta^\star} \left[\mathbf{1}\{ \operatorname{sign}(\phi[k][h][j]) \neq \operatorname{sign}(\theta^\star_h[j]) \} \right] \right)\right).
	\end{align*}
	
	Similarly, we define the event $\mathcal{E}_{k,h,j}(\theta^\star) = \{\operatorname{sign}(\phi[k][h][j]) \neq \operatorname{sign}(\theta^\star_h[j]) \}$ for any $k \in [K]$, $h \in [H]$, $j \in [d]$ and $\theta^\star_h \in \{-\Delta, \Delta\}^d$, we have  
	\begin{align*}
		2 \sum_{\theta^\star} \mathbb{E}_{\theta^\star} \left[{\rm Regret}(K)\right]  =& \frac{2 \Delta}{\sqrt{d}} \cdot \sum_{\theta^\star} \sum_{h=1}^H \sum_{j=1}^d   \left( K-  \sum_{k=1}^K \left( \mathbb{P}_{\theta^\star_{-hj}}(\mathcal{E}_{k,h,j}(\theta^\star) )- \mathbb{P}_{\theta^\star} (\mathcal{E}_{k,h,j}(\theta^\star)) \right)\right) \\
		\geq& \frac{2 \Delta}{\sqrt{d}} \cdot \sum_{\theta^\star} \sum_{h=1}^H \sum_{j=1}^d   \left( K-  K \sqrt{ \frac{1}{2}{\rm KL} \left(  \mathbb{P}_{\theta^\star}, \mathbb{P}_{\theta^\star_{-hj}} \right) } \right),
	\end{align*}
	
	Let $\mathcal{H}_{k-1}$ denote a sigma algebra w.r.t. the history before the $k$-th episode, $\mathcal{H}_{k-1} = \sigma(\{\tau^0_1, \tau^1_1, o_1, \ldots, \tau^0_{k-1}, \tau^1_{k-1}, o_{k-1}\})$. By the chain rule of relative entropy and the Bernoulli distribution of reward, we can further decompose ${\rm KL} \left(  \mathbb{P}_{\theta^\star}, \mathbb{P}_{\theta^\star_{-hj}} \right) $ as 
	\begin{align*}
		{\rm KL} \left(  \mathbb{P}_{\theta^\star}, \mathbb{P}_{\theta^\star_{-hj}} \right)  =& \sum_{k=1}^K \mathbb{E}_{\theta^\star} \left[ {\rm KL}(\mathbb{P}_{\theta^\star}(o_k \mid \tau^0_k, \tau^1_k; \mathcal{H}_{k-1}), \mathbb{P}_{\theta^\star_{-hj}}(o_k \mid \tau^0_k, \tau^1_k; \mathcal{H}_{k-1})) \right] \\
		=& \sum_{k=1}^K \mathbb{E}_{\theta^\star} \left[ {\rm KL}({\rm Ber}(\sigma(\langle \phi(\tau^0_k) - \phi(\tau^1_k), \theta^\star\rangle)), {\rm Ber}(\sigma(\langle  \phi(\tau^0_k) - \phi(\tau^1_k), \theta^\star_{-hj} \rangle))) \right].
	\end{align*}
	
	It then follows from a Taylor expansion and since $\sigma'(x) \leq 1/4$,
	\begin{align*}
		&{\rm KL}({\rm Ber}\left(\sigma(\langle \phi(\tau^0_k) - \phi(\tau^1_k), \theta^\star\rangle)), {\rm Ber}(\sigma(\langle  \phi(\tau^0_k) - \phi(\tau^1_k), \theta^\star_{-hj} \rangle))\right)\\
		=& \frac{1}{8} \cdot \langle  \phi(\tau^0_k) - \phi(\tau^1_k) , \theta^\star - \theta^\star_{-hj}  \rangle^2 \leq 2 \cdot \frac{ \Delta^2 } {d}.
	\end{align*}
	
	Then we have
	\begin{align*}
		{\rm KL} \left(  \mathbb{P}_{\theta^\star}, \mathbb{P}_{\theta^\star_{-hj}} \right) \leq 2K \cdot \frac{ \Delta^2 } {d},
	\end{align*}
	and thus
	\begin{align*}
		\sum_{\theta^\star} \mathbb{E}_{\theta^\star} \left[{\rm Regret}(K)\right]  \geq \Delta \sqrt{d} KH \sum_{\theta^\star} \left( 1 - \Delta \sqrt{K/d}  \right).
	\end{align*}
	Further let $\Delta = \sqrt{d/K}/2$, which imples that $ 1 -  \Delta \sqrt{K/d}  = 1/2$. We have there exist $\theta^\star_h \in \{-\Delta,\Delta\}^d$ for any $h \in [H]$, such that
	
	\begin{align*}
		\mathbb{E}_{\theta^\star} \left[{\rm Regret}(K)\right]   \geq \Omega\left(dH\sqrt{K}\right),
	\end{align*}
	which completes the proof.
	
\end{proof}

\section{Supporting Lemmas}\label{sec:suplemma}
\begin{lemma}[Bounding the expected $L_1$-norm of reward difference] \label{lemma:l1logeexp}
	Given any RM $r \in \mathcal{G}_r$ where $\mathcal{G}_r \in \{\mathcal{G}_r^{\rm seq}, \mathcal{G}_r^{\rm tok}\}$ denote the RM class for sequence-level RM (in Eq. (\ref{eq:rmclassseq})) or token-level RM (Eq. (\ref{eq:rmclasstoken})), we formulate the preference probability as: $\mathbb{P}_r(o=1 \mid \tau^0, \tau^1) = \sigma(r(\tau^0) - r(\tau^1))$, where $\sigma(x):=1/(1+\exp(-x))$ and satisfies $\sup_{r_{\rm min} \leq x \leq r_{\rm max}} \vert 1 / \sigma'(x)\vert \leq \kappa$, we define 
	\begin{align*}
		\ell_r(o; \tau^0, \tau^1) = 
		\begin{cases}
			\frac{1}{2} \log \frac{\mathbb{P}_r(o \mid \tau^0, \tau^1)}{\mathbb{P}_{r^\star}(o \mid \tau^0, \tau^1)}, & {\rm if}\ \mathbb{P}_{r^\star}(o \mid \tau^0, \tau^1) \neq 0; \\
			0, & {\rm otherwise}.
		\end{cases}
	\end{align*}
	Under Assumption 1, we have for some trajectory generation distributions $\pi_0$ and $\pi_1$,
	\begin{align*}
		\mathbb{E}_{\tau^0 \sim \pi_0, \tau^1 \sim \pi_1}\left[ \vert (\theta^\star - \theta)^\top (\phi(\tau^0) - \phi(\tau^1)) \vert^2 \right] \leq - \kappa^2  \log \mathbb{E}_{\tau^0 \sim \pi_0, \tau^1 \sim \pi_1 \atop o \sim \mathbb{P}_{r^\star}(\cdot \mid \tau^0, \tau^1)} \exp\left( \ell_r(o; \tau^0, \tau^1) \right).
	\end{align*}
\end{lemma}

\begin{proof}
	First, we observe that for any trajectory pair $(\tau^0, \tau^1)$, and any RM parameter $\theta$, we can apply the Lagrange mean value theorem, 
	\begin{align*}
		\vert   r^\star(\tau^0) - r^\star(\tau^1) - (r(\tau^0) - r(\tau^1))  \vert 
        =& \vert \sigma^{-1}(\mathbb{P}_{r^\star}(o=1 \mid \tau^0, \tau^1)) -  \sigma^{-1}(\mathbb{P}_{r}(o=1 \mid \tau^0, \tau^1))\vert  \\
		\leq & \sup_{r_{\rm min} \leq x\leq r_{\rm max}}\left\vert\frac{1}{\sigma'(x)}\right\vert \cdot \vert  \mathbb{P}_{r^\star}(o=1 \mid \tau^0, \tau^1) -  \mathbb{P}_{r}(o=1 \mid \tau^0, \tau^1)\vert \\
		\leq& \kappa \cdot \vert  \mathbb{P}_{r^\star}(o=1 \mid \tau^0, \tau^1) -  \mathbb{P}_{r}(o=1 \mid \tau^0, \tau^1)\vert.
	\end{align*}
	
	Thus we have
	\begin{align*}
		&\vert   r^\star(\tau^0) - r^\star(\tau^1) - (r(\tau^0) - r(\tau^1))  \vert \\
		\leq& \kappa \cdot \vert  \mathbb{P}_{r^\star}(o=1 \mid \tau^0, \tau^1) - \mathbb{P}_{r}(o=1 \mid \tau^0, \tau^1) \vert \\
		=& \frac{\kappa}{2} \cdot \Big( \left\vert  \mathbb{P}_{r^\star}(o=1 \mid \tau^0, \tau^1) - \mathbb{P}_{r}(o=1 \mid x, \tau^0, \tau^1) \right\vert  + \left\vert  \mathbb{P}_{r^\star}(o=0 \mid \tau^0, \tau^1) - \mathbb{P}_{r}(o=0 \mid \tau^0, \tau^1) \right\vert \Big) \\
		=&  \kappa \cdot {\rm TV} \left(\mathbb{P}_{r^\star}(\cdot \mid \tau^0, \tau^1) - \mathbb{P}_{r}(\cdot \mid \tau^0, \tau^1)  \right),
	\end{align*}
	where the using the definition of total variation distance (denoted by ${\rm TV}(\cdot, \cdot)$).
	
	By squaring and taking expectation on both sides, we have
	\begin{align}
		\begin{split} \label{eq:lemma1eq2}
			\mathbb{E}_{\tau^0 \sim \pi_0, \tau^1 \sim \pi_1}\left[ \vert  (\theta^\star - \theta)^\top (\phi(\tau^0) - \phi(\tau^1)) \vert^2 \right] \leq \kappa^2 \cdot \mathbb{E}_{\tau^0 \sim \pi_0, \tau^1 \sim \pi_1} \left[ \left({\rm TV} \left(  \mathbb{P}_{r^\star}(\cdot \mid \tau^{0}, \tau^{1}), \mathbb{P}_{r}(\cdot \mid \tau^{0}, \tau^{1}) \right)\right)^2 \right],
		\end{split}
	\end{align}
	where $c > 0$ denotes an universal constant. 
	
	By the Cauchy–Schwarz inequality, we have
	\begin{align*}
		&\left({\rm TV}\left(  \mathbb{P}_{r^\star}(\cdot \mid \tau^{0}, \tau^{1}) - \mathbb{P}_{r}(\cdot \mid \tau^{0}, \tau^{1})  \right)\right)^2 \\
		=& \frac{1}{4} \left(  \sum_{o \in \{0,1\}} \lvert    \mathbb{P}_{r^\star}(o \mid \tau^{0}, \tau^{1}) - \mathbb{P}_{r}(o \mid \tau^{0}, \tau^{1}) \rvert  \right)^2 \\
		\leq& \frac{1}{4}  \sum_{o \in \{0,1\}} \lvert  \sqrt{\mathbb{P}_{r^\star}(o \mid x, \tau^{0}, \tau^{1})} - \sqrt{\mathbb{P}_{r}(o \mid \tau^{0}, \tau^{1})}  \rvert^2 \sum_{o \in \{0,1\}} \lvert  \sqrt{\mathbb{P}_{r^\star}(o \mid \tau^{0}, \tau^{1})} + \sqrt{\mathbb{P}_{r}(o \mid \tau^{0}, \tau^{1})}  \rvert^2  \\
		\leq& \sum_{o \in \{0,1\}} \lvert  \sqrt{\mathbb{P}_{r^\star}(o \mid \tau^{0}, \tau^{1})} - \sqrt{\mathbb{P}_{r}(o \mid \tau^{0}, \tau^{1})}  \rvert^2.
	\end{align*}
	
	Following form Eq. (\ref{eq:lemma1eq2}), we have
	\begin{align*}
		\mathbb{E}_{\tau^0 \sim \pi_0 \mid \tau^1 \sim \pi_1} \left[\left({\rm TV} \left( \mathbb{P}_{r^\star}(\cdot \mid \tau^{0}, \tau^{1}) - \mathbb{P}_{r}(\cdot \mid \tau^{0}, \tau^{1})  \right)\right)^2  \right] \leq \mathbb{E}_{\tau^0 \sim \pi_0, \tau^1 \sim \pi_1} \left[  \sum_{o \in \{0,1\}} \lvert  \sqrt{\mathbb{P}_{r^\star}(o \mid \tau^{0}, \tau^{1})} - \sqrt{\mathbb{P}_{r}(o \mid \tau^{0}, \tau^{1})}  \rvert^2    \right].
	\end{align*}
	
	Further, we have 
	\begin{align*}
		&\mathbb{E}_{\tau^0 \sim \pi_0, \tau^1 \sim \pi_1} \left[\sum_{o \in \{0,1\}} \lvert  \sqrt{\mathbb{P}_{r^\star}(o \mid \tau^{0}, \tau^{1})} - \sqrt{\mathbb{P}_{r}(o \mid \tau^{0}, \tau^{1})}  \rvert^2 \right] \\
		=& \mathbb{E}_{\tau^0 \sim \pi_0, \tau^1 \sim \pi_1} \left[ \sum_{o \in \{0,1\}}  \mathbb{P}_{r^\star}(o \mid \tau^{0}, \tau^{1}) + \mathbb{P}_{r}(o \mid \tau^{0}, \tau^{1}) - 2 \sqrt{\mathbb{P}_{r^\star}(o \mid \tau^{0}, \tau^{1})  \mathbb{P}_{r}(o \mid  \tau^{0}, \tau^{1})} \right] \\
		=&  2- 2 \mathbb{E}_{\tau^0 \sim \pi_0, \tau^1 \sim \pi_1 \atop o \sim \mathbb{P}_{r^\star}(o \mid \tau^0, \tau^1)} \left[ \sqrt{\frac{\mathbb{P}_{r}(o \mid \tau^{0}, \tau^{1}) }{\mathbb{P}_{r^\star}(o \mid \tau^{0}, \tau^{1}) }}  \right] \\
		\leq& -2 \log \mathbb{E}_{\tau^0 \sim \pi_0, \tau^1 \sim \pi_1 \atop o \sim \mathbb{P}_{r^\star}(o \mid \tau^0, \tau^1)} \left[ \sqrt{\frac{\mathbb{P}_{r}(o \mid  \tau^{0}, \tau^{1}) }{\mathbb{P}_{r^\star}(o \mid \tau^{0}, \tau^{1}) }}  \right] \\
		=& -2 \log \mathbb{E}_{\tau^0 \sim \pi_0, \tau^1 \sim \pi_1 \atop o \sim \mathbb{P}_{r^\star}(o \mid \tau^0, \tau^1)} \left[ \exp \left( \frac{1}{2} \log \frac{\mathbb{P}_{r}(o \mid \tau^{0}, \tau^{1}) }{\mathbb{P}_{r^\star}(o \mid \tau^{0}, \tau^{1}) } \right) \right],
	\end{align*}
	where the inequality follows from the fact that $1-x \leq -\log x$.
	
	Combining the above inequalities, we complete the proof of Lemma \ref{lemma:l1logeexp}.
\end{proof}

\begin{definition}[$\epsilon$-bracketing number of sequence-level preference objectives] \label{def:bracketnumseq}
	Given a reward model $r \in \mathcal{G}_r^{\rm seq}$ defined in Eq. (\ref{eq:rmclassseq}), we define 
	\begin{align}
		\ell_r(o; \tau^0, \tau^1) = 
		\frac{1}{2} \log \frac{\mathbb{P}_{r}(o \mid \tau^0, \tau^1)}{\mathbb{P}_{r^\star}(o \mid \tau^0, \tau^1)}
	\end{align}
	as the preference objective w.r.t. the sequence-level RM $r^\star$. A pair $(\ell^1, \ell^2)$ is called an \emph{$\epsilon$-bracket} if it satisfies:
	\begin{itemize}
		\item $\ell^1(o; \tau^0, \tau^1) \leq \ell^2(o; \tau^0, \tau^1)$ for all preference samples $(o; \tau^0, \tau^1)$;
		\item The $L_\infty$-distance between the two functions is bounded as $\left\| \ell^1 - \ell^2 \right\|_\infty \leq \epsilon$.
	\end{itemize}
	The \emph{$\epsilon$-bracketing number} of $\mathcal{L}_{\rm seq}=\{\ell_r: r \in \mathcal{G}_r^{\rm seq} \}$ denoted by $\mathcal{N}_{[]}(\epsilon, \mathcal{L}_{\rm seq}, L_{\infty})$, is the minimal number of such brackets $\{(\ell^1_n, \ell^2_n)\}_{n=1}^N$ required to cover $\mathcal{L}_r$, in the sense that for every $\ell \in \mathcal{L}_{\rm seq}$, there exists an index $n \in [N]$ such that $\ell^1_n(o; \tau^0, \tau^1) \leq \ell(o; \tau^0, \tau^1) \leq \ell^2_n(o; \tau^0, \tau^1), \quad \forall (o; \tau^0, \tau^1)$.
\end{definition}

\begin{proposition}[Bounding bracketing number of sequence-level preference objectives] \label{prop:bracketing}
	Under Assumption~\ref{asp:linearboundseq}, for any $\epsilon \leq 1$, the $\epsilon$-bracketing number of the sequence-level preference objective class $\mathcal{L}_{\rm seq}$ with respect to the $L_\infty$ norm satisfies: $\log \mathcal{N}_{[]}(\epsilon, \mathcal{L}_{\rm seq}, L_{\infty}) \leq \mathcal{O} \left( d \log \frac{B}{\epsilon} \right)$.
\end{proposition}

\begin{proof}
	We first derivate the Lipschitz continuity of $\log \mathbb{P}_{r}$.
	
	For fixed \(o, \tau^0, \tau^1\), consider two cases:
	\begin{itemize}
		\item If \(o = 1\), then \(\log P_{r}(o=1 \mid \tau^0, \tau^1) = \log \sigma(\theta^\top ( \phi(\tau^0) - \phi(\tau^1)))\).
		\item If \(o = 0\), then \(\log P_{r}(o=0 \mid \tau^0, \tau^1) = \log \sigma(-\theta^\top ( \phi(\tau^0) - \phi(\tau^1)))\).
	\end{itemize}
	Let $x = \theta^\top ( \phi(\tau^0) - \phi(\tau^1))$ denote the reward difference. The derivative of $\log \sigma(x)$ is $1 - \sigma(x)$, whose absolute value is bounded by 1. Similarly, $\log \sigma(-x)$ has derivative $-\sigma(x)$, with absolute value bounded by 1. Thus, both are Lipschitz continuous with constant 1. Hence, let $r_1 \in \mathcal{G}_r$ and $r_2 \in \mathcal{G}_r$ denote the reward models parameterized by $\theta_1$ and $\theta_2$ respectively,
	\begin{align*}
		\left| \log \mathbb{P}_{r_{2}} - \log \mathbb{P}_{r_{1}} \right| \leq \left| (\theta_2 - \theta_1)^\top ( \phi(\tau^0) - \phi(\tau^1)) \right|.
	\end{align*}
	By Cauchy-Schwarz inequality,
	\begin{align*}
		\left| (\theta_2 - \theta_1)^\top (\phi(\tau^0) - \phi(\tau^1)) \right| \leq \|\theta_2 - \theta_1\|_2 \| ( \phi(\tau^0) - \phi(\tau^1))\|_2 \leq 2 \|\theta_2 - \theta_1\|_2.
	\end{align*}
	Therefore,
	\begin{align*}
		\left\Vert  \ell^{1} - \ell^{2} \right\Vert_\infty = \frac{1}{2}\left\Vert \log \mathbb{P}_{r_{2}} - \log \mathbb{P}_{r_{1}} \right\Vert_\infty \leq  \|\theta_2 - \theta_1\|_2.
	\end{align*}
	
	To form an $\epsilon$-bracket $(\ell^1, \ell^2)$, such that $\|\ell^1- \ell^2 \|_\infty \leq \epsilon$, for every objective function. Choose a set of parameters $\{\theta_i\}_{i=1}^N$ such that for any RM $r_\theta$ parameterized by $\theta$, there exists some RM $r_{\theta_i}$ parameterized with $\theta_i$ such that $\|\theta - \theta_i\|_2 \leq \frac{\epsilon}{2}$. Then define
	\begin{align*}
		\ell^1 = \ell_{r_{\theta_i}} - \frac{\epsilon}{2}, \quad \ell^2 = \ell_{r_{\theta_i}} + \frac{\epsilon}{2}.
	\end{align*}
	Clearly, $\lVert \ell^1  - \ell^2 \rVert_\infty = \epsilon$. Moreover, for any $\theta$, if $\|\theta - \theta_i\|_2 \leq \frac{\epsilon}{2}$, then by the Lipschitz property,
	\begin{align*}
		\left\Vert \ell_{r_\theta} - \ell_{r_{\theta_i}} \right\Vert_\infty \leq \|\theta - \theta_i\|_2 \leq \frac{\epsilon}{2},
	\end{align*}
	so $\ell^1 \leq \ell_{r_\theta} \leq \ell^2$. Thus, the brackets $(\ell^1, \ell^2)$ cover the function class.
	
	We now bound the number of parameters \(\theta_i\) needed using the covering number $\mathcal{N}(\frac{\epsilon}{2}, \Theta, \|\cdot\|_2)$ of the set $\Theta = \{\theta \in \mathbb{R}^d: \|\theta\|_2 \leq B\}$ under the \(\ell_2\) by Propositon \ref{pro:covervector}:
	\begin{align}
		\mathcal{N}\left(\frac{\epsilon}{2}, \Theta, \|\cdot\|_2\right) \leq \left(1 + \frac{4B}{\epsilon}\right)^d.
	\end{align}
	Therefore, the \(\epsilon\)-bracketing number is at most $\left(1 + \frac{4B}{\epsilon}\right)^d$.
	This completes the proof.
\end{proof}

\begin{definition}[$\epsilon$-bracketing number for token-level preference objectives] \label{def:bracketnumaction} \label{def:bracketnumtoken}
	Given the reward model $r \in \mathcal{G}_r^{\rm tok}$ defined in Eq. (\ref{eq:rmclasstoken}), we define 
	\begin{align}
		\ell_{r}(o; \tau^0, \tau^1) = 
		\frac{1}{2} \log \frac{\mathbb{P}_{r}(o \mid \tau^0, \tau^1)}{\mathbb{P}_{r^\star}(o \mid \tau^0, \tau^1)}
	\end{align}
	as the preference objective w.r.t. the token-level ground-truth RM $r^\star$. A pair $(\ell^1, \ell^2)$ is called an \emph{$\epsilon$-bracket} if it satisfies:
	\begin{itemize}
		\item $\ell^1(o; \tau^0, \tau^1) \leq \ell^2(o; \tau^0, \tau^1)$ for all preference samples $(o; \tau^0, \tau^1)$;
		\item The $L_\infty$-distance between the two functions is bounded as $\left\| \ell^1 - \ell^2 \right\|_\infty \leq \epsilon$.
	\end{itemize}
	The \emph{$\epsilon$-bracketing number} of $\mathcal{L}_{\rm tok}=\{\ell_{r}: r = \{r_h\}_{h=1}^H \in \mathcal{G}_r^{\rm tok} \}$ denoted by $\mathcal{N}_{[]}(\epsilon, \mathcal{L}_{\rm tok}, L_{\infty})$, is the minimal number of such brackets $\{(\ell^1_n, \ell^2_n)\}_{n=1}^N$ required to cover $\mathcal{L}_{\rm tok}$, in the sense that for every $\ell \in \mathcal{L}_{\rm tok}$, there exists an index $n \in [N]$ such that $\ell^1_n(o; \tau^0, \tau^1) \leq \ell(o; \tau^0, \tau^1) \leq \ell^2_n(o; \tau^0, \tau^1), \quad \forall (o; \tau^0, \tau^1)$.
\end{definition}

\begin{proposition}[Bounding bracketing number of token-level preference objectives] \label{prop:bracketingaction}
	Under Assumption~\ref{asp:linearboundtoken}, for any $\epsilon \leq 1$, the $\epsilon$-bracketing number of the token-level preference objective class $\mathcal{L}_{r}$ with respect to the $L_\infty$ norm satisfies: $\log \mathcal{N}_{[]}(\epsilon, \mathcal{L}_{\rm tok}, L_{\infty}) \leq \mathcal{O} \left( dH \log \frac{B\sqrt{H}}{\epsilon}\right)$.
\end{proposition}

\begin{proof}
	Using a similar procedure as in the proof of Proposition \ref{prop:bracketing}, we can derivate the Lipschitz continulity:
	\begin{align*}
		\Vert  \ell_{r_{{\theta}_2}} - \ell_{r_{{\theta}_1}} \Vert \leq \Vert {\theta}_2 - {\theta}_1 \Vert_2.
	\end{align*}
	
	We form a $\epsilon$-bracket $(\ell^1, \ell^2)=(\ell_{r_{{\theta}_i}} - \frac{\epsilon}{2}, \ell_{r_{{\theta}_i}} + \frac{\epsilon}{2})$ which satisfies $\Vert \ell^1 - \ell^2\Vert_\infty \leq \epsilon$. Such that for every ${\theta} \in \Theta^H$ where $\Theta = \{\theta \in \mathbb{R}^d: \Vert \theta \Vert \leq B\}$, we choose a parameters ${\theta}_i \in \mathbb{R}^{dH}$ from a $\frac{\epsilon}{2}$-covering set with $\Vert {\theta} - {\theta}_i \Vert_2 \leq \frac{\epsilon}{2}$. Which imples that $\ell^1 \leq \ell_{r_{\theta}} \leq \ell^2$. Thus, we can bound the number of $\epsilon$-brackets used to cover $\Theta^H$ using the covering number $\mathcal{N}(\frac{\epsilon}{2}, \Theta^H, \Vert \cdot \Vert_2)$ by Proposition \ref{pro:covercarvector}:
	\begin{align*}
		\mathcal{N}_{[]}(\epsilon, \mathcal{L}_{\rm tok}, L_{\infty}) = \mathcal{N}\left(\frac{\epsilon}{2}, \Theta^H, \Vert \cdot \Vert_2\right) \leq \left(1+\frac{4B\sqrt{H}}{\epsilon}\right)^{dH},
	\end{align*}
	which completes the proof.
\end{proof}

\begin{lemma}[Martingale exponential inequality (Theorem 13.2 of \cite{zhang2023mathematical})]
	Let $\{ \xi_s \}_{s=1}^\infty$ be a sequence of real-valued random variables adapted to filtration $\{\mathcal{F}_s \}$, It holds with probability $1-\delta$ such that for any $k \geq 1$,
	\begin{align}
		- \sum_{s=1}^k \log \mathbb{E} \left[ \exp \left( \xi_s \right) \mid \mathcal{F}_{s-1}  \right] \leq - \sum_{s=1}^k \xi_s + \log(1/\delta).
	\end{align}
\end{lemma}

\begin{lemma}[Uniform convergence for preference with bracketing number] \label{lemma:unformart} 
	Denote $\mathcal{L} \in \{\mathcal{L}_{\rm seq}, \mathcal{L}_{\rm tok} \}$, where $\mathcal{L}_{\rm seq} = \{\ell_r: r \in \mathcal{G}^{\rm seq}_r\}$ and $\mathcal{L}_{\rm tok}=\{\ell_r: r \in \mathcal{G}^{\rm tok}_r\}$ are the sequence-level/token-level preference objective class, whose $1/k$-bracketing numbers w.r.t. the $L_\infty$-norm are defined as $\mathcal{N}_{[]}(1/k, \mathcal{L}, L_\infty)$ as defined in Definitions \ref{def:bracketnumseq} \& \ref{def:bracketnumtoken}, respectively. Then for any $\delta \in (0,1]$, we have with probability at least $1-\delta$: $\forall \ell \in \mathcal{L}$,
	\begin{align}
		- \sum_{s=1}^{k} \log \mathbb{E}_{\tau^0 \sim \hat{\pi}^s, \tau^1 \sim \pi_{\rm ref}^s \atop o \sim \mathbb{P}_{r^\star}(\cdot \mid, \tau^0, \tau^1)} \exp\left(\ell(o; \tau^0, \tau^1) \right) \leq c \cdot \log \frac{\mathcal{N}_{[]}(1/k, \mathcal{L}, L_\infty)}{\delta},
	\end{align}
	with a constant $c>0$.
\end{lemma}
\begin{proof}
	
	Applying the above lemma to a sequence of random variables $\{ \ell(o_s; \tau^0_{s}, \tau^1_{s}) \}_{s=1}^\infty$ along with the filtration $\{ \mathcal{F}_s \}$ with $\mathcal{F}_s$ given by the $\sigma$-algebra of $\{ (o_i; \tau^0_{i}, \tau^1_{i}) \}_{i=1}^s$, we conclude that given a RM $r$, it holds with probability $1-\delta$,
	\begin{align*} \label{eq:martei}
		\begin{split}
			- \sum_{s=1}^{k} \log \mathbb{E}_{\tau^0 \sim \hat{\pi}^s, \tau^1 \sim \pi_{\rm ref}^s \atop o \sim p_{r^\star}(\cdot \mid \tau^0, \tau^1)} \exp\left( \ell_r(o; \tau^0, \tau^1) \right) = - \sum_{s=1}^{k} \log \mathbb{E} \left[ \exp\left( \ell_r(o_s; \tau^0_s, \tau^1_s) \right) \mid \mathcal{F}_{s-1} \right] \leq -  \sum_{s=1}^{k} \ell_r(o_s; \tau^0_{s}, \tau^1_{s}) + \log(1/\delta).
		\end{split}
	\end{align*}

	Let $\widetilde{\mathcal{L}}$ denote a minimal $1/k$-bracket set of the preference objective class $\mathcal{L}$ with a size of $\mathcal{N}_{[]}(1/k, \mathcal{L}, L_\infty)$. Given an online preference set $\mathcal{D}_k^{\rm pref} = \{(o_s; \tau^0_s, \tau^1_s)\}_{s=1}^k$, where $\tau^0_s \sim \hat{\pi}^s$, $\tau^1_s \sim \pi_{\rm ref}^s$, $o_s \sim \mathbb{P}_{r^\star}(\cdot \mid \tau^0_{s}, \tau^1_{s})$, let
	\begin{align*}
		u(\ell, \mathcal{D}_k^{\rm pref}) := \sum_{s=1}^{k} \ell(o_s; \tau^0_{s}, \tau^1_{s}) - \sum_{s=1}^{k} \log \mathbb{E}_{\tau^0 \sim \hat{\pi}^s, \tau^1 \sim \pi_{\rm ref}^s \atop o \sim \mathbb{P}_{r^\star}(\cdot \mid \tau^0, \tau^1)} \exp\left( \ell(o; \tau^0, \tau^1) \right),
	\end{align*}
	then there exists $(\ell^1, \ell^2) \in \widetilde{\mathcal{L}}$ such that
	\begin{align*}
		\sup_{\ell \in \mathcal{L}} \left\{  u(\ell, \mathcal{D}_k^{\rm pref}) \right\} \leq \sup_{\ell \in \mathcal{L}}  \left\{  u(\ell, \mathcal{D}_k^{\rm pref}) - u(\ell^1, \mathcal{D}_k^{\rm pref}) + u(\ell^1, \mathcal{D}_k^{\rm pref}) \right\}.
	\end{align*}
	
	Following from
	\begin{align*}
		&u(\ell, \mathcal{D}^{\rm pref}_k) -  u(\ell^1, \mathcal{D}^{\rm pref}_{k}) \\
		=& \sum_{s=1}^k \ell(o_s; \tau^0_{s}, \tau^1_{s}) - \ell^1(o_s; \tau^0_{s}, \tau^1_{s}) + \sum_{s=1}^k  \log \mathbb{E}_{\tau^0 \sim \hat{\pi}^s, \tau^1 \sim \pi_{\rm ref}^s \atop o \sim \mathbb{P}_{r^\star}(\cdot \mid \tau^0, \tau^1)} \exp\left( \ell^1(o; \tau^0, \tau^1) \right)  -   \log \mathbb{E}_{\tau^0 \sim \hat{\pi}^s, \tau^1 \sim \pi_{\rm ref}^s \atop o \sim \mathbb{P}_{r^\star}(\cdot \mid \tau^0, \tau^1)} \exp\left( \ell(o; \tau^0, \tau^1) \right)\\
		\leq& 2k \Vert \ell - \ell^1 \Vert_\infty \leq 2k \Vert \ell^2 - \ell^1 \Vert_\infty,
	\end{align*}
	then we have
	\begin{align*}
		\sup_{\ell \in \mathcal{L}} \left\{  u(\ell, \mathcal{D}_k^{\rm pref}) \right\} \leq& \sup_{(\ell^1, \ell^2) \in \widetilde{\mathcal{L}}} \left\{ 2k \Vert \ell^2 - \ell^1 \Vert_\infty + u(\ell^1, \mathcal{D}_k^{\rm pref}) \right\} \\
		\leq& 2 +  \sup_{(\ell^1, \ell^2) \in \widetilde{\mathcal{L}}}\left\{ \sum_{s=1}^{k} \ell^1(o_s; \tau^0_{s}, \tau^1_{s}) - \sum_{s=1}^{k} \log \mathbb{E}_{\tau^0 \sim \hat{\pi}^s, \tau^1 \sim \pi_{\rm ref}^s \atop o \sim \mathbb{P}_{r^\star}(\cdot \mid \tau^0, \tau^1)} \exp\left( \ell^1(o; \tau^0, \tau^1) \right) \right\},
	\end{align*}
	where the second inequality follows that $\widetilde{\mathcal{L}}$ is the $1/k$ bracketing set of $\mathcal{L}$. 
	
	Using Lemma \ref{lemma:unformart} with an union bound, we have for any $\delta \in (0,1]$, with probability at least $1 - \delta$,
	\begin{align*}
		\sup_{(\ell^1, \ell^2) \in \widetilde{\mathcal{L}}}\left\{ \sum_{s=1}^{k} \ell^1(o_s; \tau^0_{s}, \tau^1_{s}) - \sum_{s=1}^{k} \log \mathbb{E}_{\tau^0 \sim \hat{\pi}^s, \tau^1 \sim \pi_{\rm ref}^s \atop o \sim \mathbb{P}_{r^\star}(\cdot \mid \tau^0, \tau^1)} \exp\left( \ell^1(o; \tau^0, \tau^1) \right) \right\} \leq \log \frac{\mathcal{N}_{[]}(1/k, \mathcal{L}, L_\infty)}{\delta},
	\end{align*}
	for some constant $c>0$.
	
	Then, we have for any $\delta \in (0,1]$, with probability at least $1 - \delta$, $\forall \ell \in \mathcal{L}$:
	\begin{align*}
		& - \sum_{s=1}^{k} \log \mathbb{E}_{\tau^0 \sim \hat{\pi}^s, \tau^1 \sim \pi_{\rm ref}^s \atop o \sim \mathbb{P}_{r^\star}(\cdot \mid \tau^0, \tau^1)} \exp\left( \ell(o; \tau^0, \tau^1) \right) \leq - \sum_{s=1}^{k} \ell(o_s; \tau^0_{s}, \tau^1_{s})  + c \log \frac{\mathcal{N}_{[]}(1/k, \mathcal{L}, L_\infty)}{\delta}   \\
		=&  \sum_{s=1}^{k} \log \frac{\mathbb{P}_{r^\star}(o_s \mid \tau^0_{s}, \tau^1_{s})}{\mathbb{P}_{r}(o_s \mid \tau^0_{s}, \tau^1_{s})} +  c \log \frac{\mathcal{N}_{[]}(1/k, \mathcal{L}, L_\infty)}{\delta}  \leq  \sum_{s=1}^{k} \log \frac{\mathbb{P}_{\hat{r}_k}(o_s\mid \tau^0_{s}, \tau^1_{s})}{\mathbb{P}_{r}(o_s\mid \tau^0_{s}, \tau^1_{s})}  + c \log \frac{\mathcal{N}_{[]}(1/k, \mathcal{L}, L_\infty)}{\delta}  \\
		\leq& \zeta_k +c \log \frac{\mathcal{N}_{[]}(1/k, \mathcal{L}, L_\infty)}{\delta}  \leq c \log \frac{\mathcal{N}_{[]}(1/k, \mathcal{L}, L_\infty)}{\delta},
	\end{align*}
	where the second inequality follows from $\hat{r}_k = \mathop{\arg\max}_{r \in \mathcal{G}_r} \sum_{s=1}^{k} \mathbb{P}_{r} (o_s\mid \tau^0_{s}, \tau^1_{s})$, the last step follows by setting $\zeta_k = \mathcal{O} \left( \log (\mathcal{N}_{[]}(1/k, \mathcal{L}, L_\infty)/\delta) \right)$. 
	This completes the proof of Lemma \ref{lemma:unformart}.
\end{proof}

\begin{lemma}[Concentration of inverse covariances \cite{zanette2021cautiously}] \label{lemma:inversecov}
	Let $\{\phi_s\}_{s=1}^{\infty}$ be a sequence of $d$-dimensional random vectors adapted to the filtration $\{\mathcal{F}_s\}$. Assume $\Vert \phi_s\Vert_2 \leq A$ and define $\Sigma_k = \lambda I + \sum_{s=1}^{k-1} \mathbb{E}[ \phi_s \phi_s^\top \mid \mathcal{F}_{s-1}]$.  If $\lambda \geq \Omega(d\log(k/\delta))$, then we have for any $\delta \in (0,1]$ and any $k \geq 1$, with probability at least $1-\delta$,
	\begin{align*}
		\frac{3}{5} (n\Sigma_k + \lambda I)^{-1} \preceq \left( \sum_{s=1}^{k-1} \phi_i \phi_i^\top + \lambda I \right)^{-1} \preceq 3 (n\Sigma_k + \lambda I)^{-1} 
	\end{align*} 
\end{lemma}

\begin{lemma}[Performance of preference optimization] \label{lemma:mleper}
	Given a dataset $\{(o_s, \tau^0_s, \tau^1_s)\}_{s=1}^k$, we have for all reward model $r$, with probability at least $1-\delta$, 
	\begin{align*}
		\sum_{s=1}^{k} \log \frac{\mathbb{P}_{r}(o_s \mid \tau^0_s, \tau^1_s)}{\mathbb{P}_{r^\star}(o_s\mid \tau^0_s, \tau^1_s)} \leq c\cdot \log(\mathcal{N}_{[]}(1/n,\mathcal{L}, L_\infty)/\delta),
	\end{align*}
	with a constant $c>0$, $\mathcal{N}_{[]}(1/k,\mathcal{L}, L_\infty)$ denotes a $1/k$-bracketing number of $\mathcal{L}$ as defined in Lemma \ref{lemma:unformart}.
\end{lemma}

\begin{proof}
	Using the definiton of random variables $\{ \ell_r(o_s; \tau^0_{s}, \tau^1_{s}) \}_{s=1}^\infty$ along with the filtration $\{ \mathcal{F}_s \}$ in Lemma \ref{lemma:unformart}, for each $k \in [K]$, we have
	\begin{align*}
		&\mathbb{E}\left[ \exp\left( \sum_{s=1}^{k} \log \frac{\mathbb{P}_r(o_s\mid \tau^0_s, \tau^1_s)}{\mathbb{P}_{r^\star}(o_s\mid \tau^0_s, \tau^1_s)}   \right)  \right] \\
		=& \mathbb{E}\left[ \exp\left( \sum_{s=1}^{k-1} \log \frac{\mathbb{P}_r(o_s\mid \tau^0_s, \tau^1_s)}{\mathbb{P}_{r^\star}(o_s\mid  \tau^0_s, \tau^1_s)}   \right) \cdot \mathbb{E}\left[ \exp\left( \log \frac{\mathbb{P}_r(o_k\mid \tau^0_k, \tau^1_k)}{\mathbb{P}_{r^\star}(o_k \mid \tau^0_k, \tau^1_k)} \right) \mid \mathcal{F}_{k-1} \right] \right] \\
		=& \mathbb{E}\left[ \exp\left( \sum_{s=1}^{k-1} \log \frac{\mathbb{P}_r(o_s\mid \tau^0_s, \tau^1_s)}{\mathbb{P}_{r^\star}(o_s\mid \tau^0_s, \tau^1_s)}   \right) \cdot \mathbb{E}\left[ \frac{\mathbb{P}_r(o_k\mid \tau^0_k, \tau^1_k)}{\mathbb{P}_{r^\star}(o_k \mid \tau^0_k, \tau^1_k)} \mid \mathcal{F}_{k-1} \right] \right] \\
		\leq& \mathbb{E}\left[ \exp\left( \sum_{s=1}^{k-1} \log \frac{\mathbb{P}_r(o_s\mid \tau^0_s, \tau^1_s)}{\mathbb{P}_{r^\star}(o_s \mid \tau^0_s, \tau^1_s)}   \right) \cdot \left(1+\frac{1}{K}\right) \right] \leq \ldots \leq e.
	\end{align*}
	
	Then by Markov’s inequality, we have for any $\delta \in (0,1]$,
	\begin{align*}
		\mathbb{P} \left( \sum_{s=1}^{k} \log \frac{\mathbb{P}_r(o_s \mid \tau^0_s, \tau^1_s)}{\mathbb{P}_{r^\star}(o_s\mid \tau^0_s, \tau^1_s)} \geq \log(1/\delta)  \right) \leq e\delta.
	\end{align*}
	
	Using an union bound on the function class with the bracketing number technique, and rescaling $\delta$ we have
	\begin{align*}
		\mathbb{P} \left( \sum_{s=1}^{k} \log \frac{\mathbb{P}_r(o_s\mid \tau^0_s, \tau^1_s)}{\mathbb{P}_{r^\star}(o_s\mid \tau^0_s, \tau^1_s)} \geq c \cdot \log(\mathcal{N}_{[]}(1/k,\mathcal{L}, L_\infty)/\delta)  \right) \leq \delta,
	\end{align*}
	which completes the proof.
\end{proof}	

\section{Technical Lemmas}
\begin{proposition}[Covering number of the bounded vector space] \label{pro:covervector}
	The covering number $\mathcal{N}(\frac{\epsilon}{2}, \Theta, \|\cdot\|_2)$ of the set $\Theta = \{\theta \in \mathbb{R}^d: \|\theta\|_2 \leq B\}$ under the \(\ell_2\) norm can be upper bounded by
	\begin{align*}
		\mathcal{N}\left(\frac{\epsilon}{2}, \Theta, \|\cdot\|_2\right) \leq \left(1 + \frac{4B}{\epsilon}\right)^d.
	\end{align*}
\end{proposition}

\begin{proposition}[Covering number of Cartesian producting vector space]\label{pro:covercarvector}
	We define the Cartesian product of bounded vector space as: $\Theta^H = \{\theta=(\theta_1, \ldots, \theta_H): \theta_h \in \Theta, h \in [H]\}$, where $\Theta = \{\theta \in \mathbb{R}^d: \|\theta\|_2 \leq B\}$. The covering number of $\Theta^H$ can be upper bounded by
	\begin{align*}
		\mathcal{N}\left(\frac{\epsilon}{2}, \Theta^H, \Vert \cdot \Vert_2\right) \leq \left(1+\frac{4B\sqrt{H}}{\epsilon}\right)^{dH}.
	\end{align*}
\end{proposition}
\begin{proof}
	We can bound the number of $\epsilon$-brackets used to cover $\Theta^H$ using the covering number $\mathcal{N}(\frac{\epsilon}{2}, \Theta^H, \Vert \cdot \Vert_2)$:
	\begin{align*}
		\mathcal{N}\left(\frac{\epsilon}{2}, \Theta^H, \Vert \cdot \Vert_2\right) \leq \prod_{h=1}^{H} \mathcal{N}\left(\frac{\epsilon}{2\sqrt{H}}, \Theta, \Vert \cdot \Vert_2\right) \leq \left(1+\frac{4B\sqrt{H}}{\epsilon}\right)^{dH},
	\end{align*}
	where the first inequality follows from the fact that for any $\theta=(\theta_1, \theta_2, \ldots, \theta_H) \in \Theta^H$, we can choose $\tilde{\theta} = (\tilde{\theta}_1, \ldots, \tilde{\theta}_H)$ where $\tilde{\theta}_h$ is in a $\frac{\epsilon}{2\sqrt{H}}$-covering set of the $h$-th $\Theta$, such that $\{\theta \}$ forms a $\frac{\epsilon}{2}$-covering set of $\Theta^H$ since $\Vert \theta - \tilde{\theta}\Vert_2 = \sqrt{\sum_{h=1}^H \Vert \theta_h - \tilde{\theta}_h \Vert_2^2} \leq \sqrt{\sum_{h=1}^{H} (\frac{\epsilon}{2\sqrt{H}})^2} = \frac{\epsilon}{2}$.
\end{proof}

\begin{lemma}[Elliptical potential lemma \cite{dani2008stochastic,abbasi2011improved}] \label{lemma:epl}
	\label{lem:elliptical-potential}
	Let $V_0=\lambda I_d$ with $\lambda>0$. For $k=1,\dots,K$ define
	\begin{align*}
	V_k = V_{k-1} + x_k x_k^\top,
	\qquad x_k\in\mathbb{R}^d.
	\end{align*}
	Then
	\begin{align*}
	\sum_{k=1}^K \min\bigl(1,\; \|x_k\|_{V_{k-1}^{-1}}^2\bigr)
	\le 2\log\frac{\det(V_K)}{\det(V_0)}.
	\end{align*}
	Moreover, if $\|x_k\|_2 \le L$ for all $k$, then
	\begin{align*}
	\sum_{k=1}^K \min\bigl(1,\; \|x_k\|_{V_{k-1}^{-1}}^2\bigr)
	\le 2d\log \Bigl(1 + \frac{K L^2}{\lambda d}\Bigr).
	\end{align*}
\end{lemma}
\end{document}